\documentclass[manuscript,acmsmall,screen,authorversion,nonacm]{acmart}

\AtBeginDocument{%
  }

\setcopyright{acmcopyright}
\copyrightyear{2023}
\acmYear{2023}
\acmDOI{XXXXXXX.XXXXXXX}

\acmConference[Conference]{Make sure to enter the correct conference title from your rights confirmation email}{January 1, 2024}{Woodstock, NY}
\acmPrice{15.00}
\acmISBN{978-1-4503-XXXX-X/18/06}

\usepackage{latex-macros}

\usepackage[suppress]{color-edits}
\usepackage{caption}
\usepackage{subfigure}
\usepackage[capitalize,noabbrev]{cleveref}
\usepackage{enumitem}
\usepackage{tikz}

\addauthor{mp}{green}
\addauthor{ch}{blue}
\addauthor{r}{red}

\usepackage{bm}

\renewcommand{\vec}[1]{\bm {#1}}

\usepackage{listings}
\begin{document}

\title[Leveraging LLMs for Collective Decision-Making]{Leveraging Large Language Models for Collective Decision-Making} \thanks{Some of the information in this document relates to pre-released content which may be subsequently modified. Microsoft makes no warranties, express or implied, with respect to the information provided here.  This document is provided “as-is”. Information and views expressed in this document, including URL and other Internet Web site references, may change without notice.  Some examples depicted herein are provided for illustration only and are fictitious. No real association or connection is intended or should be inferred.  This document does not provide you with any legal rights to any intellectual property in any Microsoft product.

© 2023 Microsoft. All rights reserved.}

\author{Marios Papachristou}
\authornote{Work done while at Microsoft.}
\email{papachristoumarios@gmail.com}
\orcid{0000-0002-1728-0729}
\affiliation{%
  \institution{Cornell University}
  \city{Ithaca}
  \state{New York}
  \country{USA}
}

\author{Longqi Yang}
\email{loy@microsoft.com}
\orcid{0000-0002-6615-8615}
\affiliation{
  \institution{Microsoft}
  \city{Redmond}
  \state{Washington}
  \country{USA}
}

\author{Chin-Chia Hsu}
\email{chinchiahsu@microsoft.com}
\orcid{0000-0003-0144-9822}
\affiliation{
  \institution{Microsoft}
  \city{Redmond}
  \state{Washington}
  \country{USA}
}

\renewcommand{\shortauthors}{Papachristou et al.}

\acmConference[CSCW '25]{The 28th ACM SIGCHI Conference on Computer-Supported Cooperative Work \& Social Computing}{November 9-13, 2024}{San Jose, Costa Rica}


\begin{abstract}
    
In various work contexts, such as meeting scheduling, collaborating, and project planning, collective decision-making is essential but often challenging due to diverse individual preferences, varying work focuses, and power dynamics among members. To address this, we propose a system leveraging Large Language Models (LLMs) to facilitate group decision-making by managing conversations and balancing preferences among individuals. Our system aims to extract individual preferences from each member's conversation with the system and suggest options that satisfy the preferences of the members. 
We specifically apply this system to corporate meeting scheduling.
We create synthetic employee profiles and simulate conversations at scale, leveraging LLMs to evaluate the system performance as a novel approach to conducting a user study. 
Our results indicate efficient coordination with reduced interactions between the members and the LLM-based system.
The system refines and improves its proposed options over time, ensuring that many of the members' individual preferences are satisfied in an equitable way. 
Finally, we conduct a survey study involving human participants to assess our system's ability to aggregate preferences and reasoning about them. Our findings show that the system exhibits strong performance in both dimensions.

\end{abstract}



\maketitle

\section{Introduction}
Collective decision-making is ubiquitous in our daily lives, spanning from small-scale event planning within a group to public discourse on government policies that require citizens’ opinions \cite{yardi2005vern,alvarez2016hootle+}. During these discussions, participants express their individual preferences and viewpoints in their natural languages, aiming to arrive at a decision that balances the participants’ individual needs. 
However, extracting preferences from these conversations is not straightforward. Furthermore, participants’ preferences or opinions are often diverse. These imply that understanding the participants’ preferences and reaching a collective decision that respects their preferences can be time-consuming and require substantial management (cf. \cite{cranshaw2017calendar}).

In particular, in work environments such as organizing meetings, collective decision-making tasks have been a productivity bottleneck \cite{koutsuv2023focus} due to their inherent diversity of preferences across members from possibly different organizations \cite{crawford2007experts,crawford2009learning,ephrati1994meet,benhassine2007agent,brzozowski2006grouptime,mok2023challenging}, and have become more complex in the hybrid work setting\cite{yang2022effects,yang2023large,yang2022future}. Besides traditional methods using calendars (e.g., Doodle, when2meet, MS Outlook, etc.), copilots such as \emph{Calendar.help} \cite{cranshaw2017calendar} have been proposed to aid with meeting scheduling and distributed decision-making (e.g., \cite{yardi2005vern, alvarez2016hootle+,cao2018attentive}) across individuals within or across organizations. However, these approaches have been tailored to specific sub-problems and cannot encompass general preferences and constraints the members may have.\footnote{See \cref{sec:related_work} for a thorough comparison with these related works.} \chedit{For example, scheduling preferences like ``avoiding back-to-back meetings" or ``preferring flexibility and variety in the schedule" are nuanced and require sophisticated handling.}

The advent of Large Language Models (LLMs) \cite{openai2023gpt,touvron2023llama2,zhao2023survey,bubeck2023sparks} presents a promising solution to alleviate these challenges associated with collective decision-making. 
LLMs have proven their efficacy in a variety of tasks that include discerning intent from a given context and generating corresponding responses in natural languages (see, e.g., \cite{openai2023gpt,touvron2023llama2,ziegler2019fine,bai2022training} and the references therein). 
They also can offer creative output across various domains (cf. \cite{franceschelli2023creativity,gero2022sparks,salvagno2023can,chopra2023conducting,copilot,lee2022evaluating,peng2023impact,nguyen2022empirical,saharia2022photorealistic,ramesh2022hierarchical}), and aid in deliberative processes \cite{small2023opportunities,argyle2023ai}.
Considering this capacity, LLMs can potentially extract individual preferences from people’s expressions in a group discussion.
Furthermore, they may even be capable of effectively aggregating the preferences, proposing a mutually agreed decision \cite{fish2023generative}.

In this paper, we leverage the power of LLMs to create a system that assists collective decision-making within a group, such as selecting a meeting time or deciding a venue for a group event. The system's main goal is to guide the group towards an option that respects the opinions of the members.
LLMs play a crucial role in this system by extracting individual preferences from the members, consolidating these preferences, and subsequently proposing options. 
Importantly, the system also utilizes the LLMs to provide the reasoning behind each suggestion, ensuring transparency and understanding among the group members.

\begin{figure}
    \centering
    \subfigure[Group discussion with assistance by LLMs\label{subfig:high_level_system}]{\includegraphics[width=0.45\textwidth]{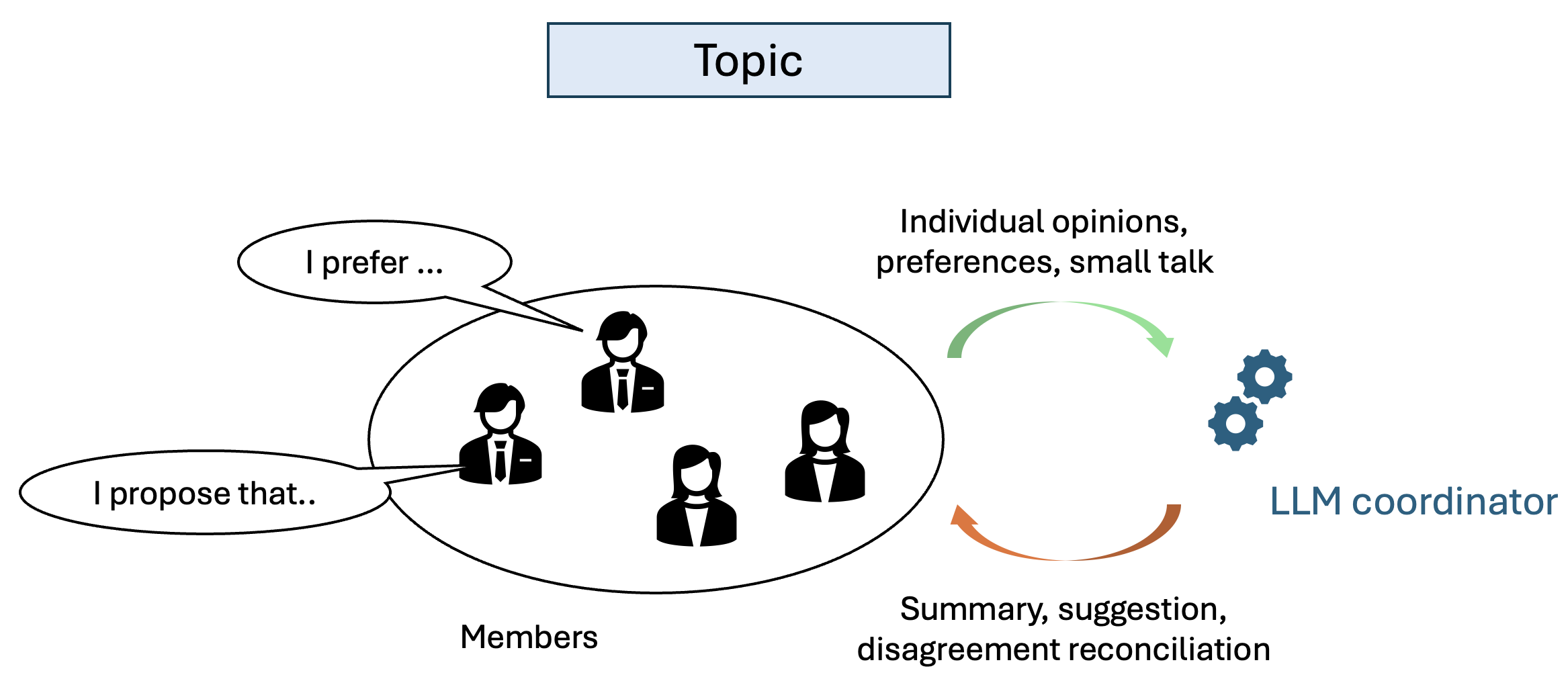}}
    \subfigure[Block diagram of the system \label{subfig:information_flow_diagram}]{\includegraphics[width=0.45\textwidth]{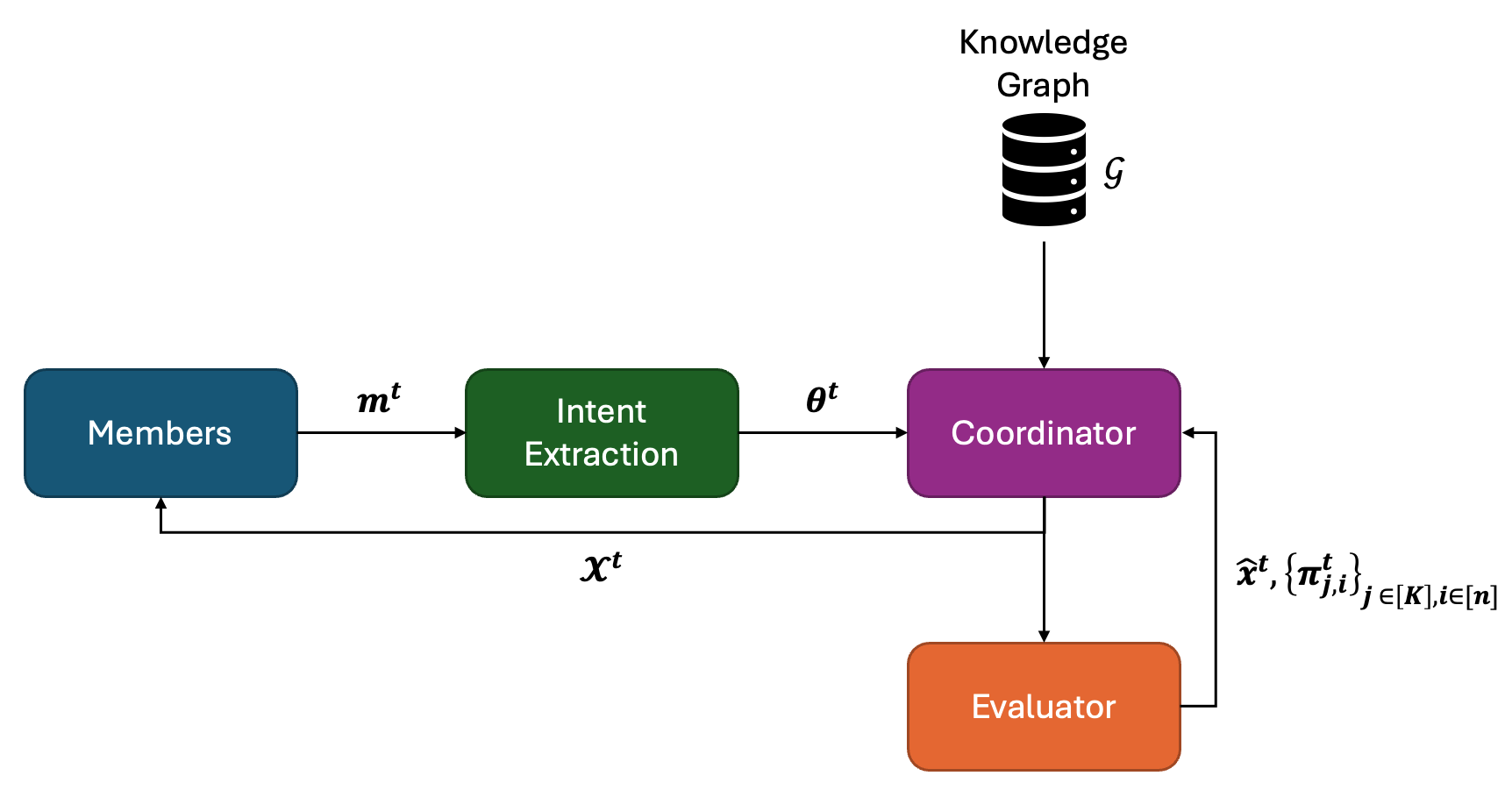}}
    \caption{Collective decision-making leveraging LLMs. (a) A high-level overview of our proposed system: The LLM-based system processes individual preferences, opinions, and small talk and suggests options that satisfy as many members as possible. (b) A block diagram of our LLM-based system as described in \cref{sec:system_design}. The members provide messages $\vec m^t$ to the intent extraction module, which distills the true preferences $\vec \theta^t$ from $\vec m^t$. The coordination module also has access to a database $\cD$. Subsequently, the coordination module (coordinator) is responsible for suggesting the option set $\cX^t = \{ x_1^t, \dots, x_K^t \}$, and the evaluation module is responsible for assigning a score $\pi_{j, i}^t$ for every member's preferences $\theta_i^t$ and every option $x_j^t$. The decision candidate is selected as the ``best'' (see \cref{eq:majority}) option out of the option set $\cX^t$ and is provided to the coordination module for round $t + 1$.}
    \label{fig:coordination_system}
\end{figure}

Specifically, our system is designed with four key components: (i) \chedit{a database}, which serves as a repository for background information about the users; (ii) an intent extraction module, tasked with discerning individual preferences from the dialogues between the system and the members; (iii) a coordination module, responsible for generating options that align with the members’ preferences; (iv) an evaluation module that assesses how well the proposed options meet the members’ preferences. \cref{fig:coordination_system} provides a comprehensive overview of our system.

To demonstrate the applicability of our system in facilitating collective decision-making, we particularly examine its performance in the context of meeting scheduling tasks.
In such meeting scenarios, the system considers not only the members’ schedule preferences but also factors like the purpose of the meeting, the members’ responsibilities, and their social connections to propose a meeting time for the group.

Our assessment consists of two parts: an evaluation of the coordination process and an assessment of reasoning and option suggestions by LLMs. 

In our first study, we conducted a large-scale experiment on a synthetic company and various meeting scenarios to measure three facets of coordination processes. These facets are the \emph{efficiency} of communication between the system and employees, the  \emph{effectiveness} of proposing options that respect the employees' preferences, and the  \emph{quality} of options in terms of fulfilling many preferences of the employees in a balanced manner. In particular, we utilized LLMs to simulate employee behavior of interacting with the system based on synthetic employee profiles (e.g., roles, schedule preferences, responsibilities) and meeting scenarios. 
This approach of using LLMs to simulate human behavior in performing tasks has been proposed by recent works such as \cite{aher2022using}, \cite{argyle2022out}, and \cite{park2023generative}.
While LLMs may not perfectly replicate human behavior in conversations, this approach offers greater control and scalability for experimentation, addressing the challenge of managing numerous meeting scheduling scenarios and discussions with human participants on a scale large enough to ensure statistical confidence.

In the second study, we recruited human participants to evaluate LLM's capacity to aggregate preferences and reasoning quality. Participants were presented with a simulated meeting scenario, including members’ schedule preferences and suggested options, along with justifications generated by the coordination module.
The participants were then tasked with evaluating the acceptability of the proposed options and their accompanying reasons. Additionally, they were asked to express their confidence level in utilizing LLMs as a tool for assisting in collective decision-making.

The findings of our large-scale study show a consistent decrease in the number of interactions between the members and the system with each coordination round. The system demonstrates effectiveness in proposing options that increasingly satisfy a greater number of members and balance their preferences over time.
Our survey results indicate that the human participants view the system as having strong capabilities for preference aggregation and reasoning. 
The participants also expressed their confidence in utilizing LLMs to coordinate members’ preferences during collective decision-making processes.

Beyond meeting scheduling, our proposed system has potential applications in various work scenarios that necessitate a balance between individual preferences, social factors, and even the infusion of creative input. For example, the system could be tailored to moderate group discussions in situations where members have conflicting opinions. Moreover, it could be designed to navigate brainstorming sessions by contributing innovative ideas and fostering new connections among members’ thoughts.

Unlike many current applications that focus on the experience of a single user interacting with LLMs,  our system emphasizes the collective experience of a group of members leveraging the capabilities of LLMs. This innovation opens up many possibilities for employing LLMs in future work environments.

\smallskip 
\noindent \textbf{Paper Organization.} The paper is organized as follows: In \cref{sec:requirements_analysis}, we perform a requirement analysis for our proposed system, analyzing responses from our participants regarding their characteristics and needs and comparing our system design with existing systems. \cref{sec:system_design} introduces the main components of our system and the collective decision-making process. In \cref{sec:case_study}, we focus on group meeting scheduling as a case study for our system. In \cref{sec:eval_quant,sec:results}, we conduct a large-scale simulation study to evaluate our system's performance, and in \cref{sec:eval_qualitative}, our study participants evaluate the system's reasoning abilities. Finally, we discuss our findings in \cref{sec:discussion} and review related literature in \cref{sec:related_work}.

\smallskip
\noindent \textbf{Supplemental Information.} 
The appendix of our paper has been uploaded as supplementary material. The Appendix details the system implementation prompts, survey design, and experimental results.

\section{Requirements Analysis} \label{sec:requirements_analysis}

In this section, we aim to understand user characteristics, usage contexts, user needs, and previous methods or systems. We performed a user study in a technology company; we recruited the employees via internal campaigns and mailing lists. The study was approved by the ethics committee within our organization (cf. \cref{sec:ethics_statement}), and the participants who completed our study were compensated with 15 US dollars.  For more information about the study design, see \cref{app:user_study}.

\subsection{Participants}
\begin{figure}[t]
    \centering
    \includegraphics[width=\textwidth]{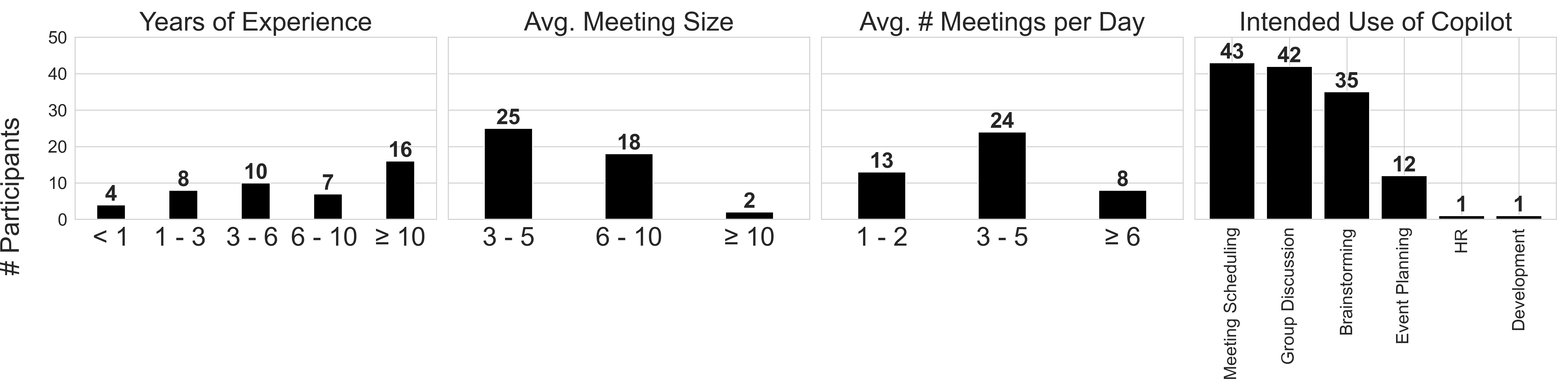}
    \caption{Participant Responses.}
    \label{fig:demographics}
\end{figure}

We received a total of 45 effective responses.
The participants held various roles (e.g., software engineer, product manager, etc.). 
Most of the participants (78\%) had at least 3 years of experience.
About half of the participants (55\%) indicated that their meetings usually had 3 to 5 attendees, and few of them had meetings with sizes larger than 10. 
Regarding the meeting frequency, 28\% of the participants had between 1 and 2 meetings per day, 54\% of them had between 3 and 5 meetings per day, and the rest had more than 6 meetings per day.
Mid-senior employees had the most meetings per day.  \cref{fig:demographics} plots the statistics.


\subsection{Usage Contexts and User Needs}

\chedit{In an organization, many decisions require input from involved members, who not only indicate their choices but also share the reasons behind them. Our LLM-based system is designed to gather members' opinions and the rationale behind their choices, ensuring that individual perspectives are considered in group recommendations rather than simply relying on a majority vote.

This is especially important when a group includes participants external to the organization, as coordinators may need extensive communication to understand and respect these participants’ preferences. Our system aims to reduce the communication workload and efficiently identify options that balance the preferences of all members.

}

The participants were asked to indicate their intended use of an LLM-based copilot for collective decision-making.
They were allowed to provide any scenarios in the free text, but we offered several potential cases: (1) meeting scheduling, (2) group discussion for distributed communication applications, (3) brainstorming for collaborative applications, and (4) event planning for workplace management. We found that meeting scheduling was considered an important use case by almost all (95\%) of the participants,  followed by group discussion (94\%), brainstorming (78\%), and event planning (27\%). Two participants also mentioned the use of HR tools and software development. 
This survey result suggested meeting scheduling as an important scenario of our proposed framework, and we will use it for our case study.

\chedit{Survey participants, who serve as administrators and frequently coordinate numerous events, shared their coordination experiences in detail. They primarily organize frequent meetings, team events, and executive gatherings, dedicating around 70\% of their weekly work time to these tasks. These events typically involve 2 to 20 participants, with about 10\% from external organizations. Their common scheduling challenges are time zone differences and often extensive communication. The administrators also indicate several decision-making factors, such as executive attendance, travel, budget, and space availability, balanced by prioritizing meeting owners' needs. 

During their coordination process, commonly used tools include email clients, team messaging applications, note-taking applications, and office suite tools (such as spreadsheets) valued for their integrated functionality. On average, the administrators propose three scheduling options. 
Although 25\% of plans may not work for everyone, attendance remains high at 90\%, and the administrators express their confidence that preferences are considered equitably. The administrators highlight additional challenges, including the need for supportive scheduling tools, educating others on scheduling protocols, and accommodating executives’ requests, emphasizing the demand for greater flexibility. This underscores a clear use case for our system, which aims to address these specific coordination and scheduling needs effectively.}

\subsection{Existing Methods or Systems}
\chedit{The surveyed administrators highlighted commonly used coordination tools, including email clients (Outlook, Gmail, Thunderbird, etc.), instant messaging apps (Teams, Slack, Google Chat, etc.), note-taking tools (Loop, Google Notes, etc.), and spreadsheet tools (Excel, Google Sheets, etc.). Additional tools like Doodle, When2Meet, and Outlook facilitate the collection of time slot preferences and help identify a majority, focusing primarily on availability statistics with minimal emphasis on participants' reasons for their choices. By contrast, our system is designed to offer greater interpretability by integrating individual preferences and rationales into its recommendations.
Given this focus, a direct comparison between our proposed system and these popular polling methods is less meaningful, as our system processes the nuanced preferences behind choices rather than treating each participant’s selection with equal weight.

Additionally, these traditional tools are less effective when some participants are external to the organization, as coordinators often cannot directly access external participants' calendars. This limitation necessitates extensive back-and-forth communication (e.g., emails), which can cause delays. Our system addresses this challenge by streamlining the process of collecting and aggregating participants' preferences, reducing communication overhead.}

A closely related work to ours is \emph{Calendar.help} \cite{cranshaw2017calendar}, which designs a meeting scheduling assistant through structured workflows. 
Specifically, users interact with \emph{Calendar.help} by email and \emph{Calendar.help} parses the workflows and schedules meetings by extracting information from the users' calendars of meetings and tasks where human-in-the-loop interventions may be needed.\footnote{For more details, \emph{Calendar.help} consists of microtask execution layers that are based on a combination of machine learning, natural language processing, and heuristic-based components to extract information about the meeting request, details about the attendees, the subject, the time, and the location of the meeting and coordinate the meeting time with the attendees. When the automatic microtasks fail, \emph{Calendar.help} relies on human-in-the-loop interventions to resolve the scheduling issues. }

Our system shares several features with \emph{Calendar.help}, but is also significantly different:
Specifically, we design four modules to extract individual preferences from dialogues and suggest meeting times, as will be detailed in \cref{sec:system_design}.
Our system is also dynamic in the sense that it aims to improve its proposed meeting plan through several rounds of conversation with the group members.
On the other hand, \emph{Calendar.help} supports a limited amount of tasks and uses a complicated architecture, unable to capture diverse free-text preferences.
This work also took significant time (18 months) to complete its user study. 
By contrast, our work will introduce some innovations: Our model can process free-text preferences. Its simple architecture of modules that are implemented using prompts allows for quick adaptation to a variety of tasks (e.g., event coordination) beyond meeting scheduling.
Moreover, we will adopt an innovative approach of conducting a large-scale user study using LLM-simulated users, therefore reducing costs and facilitating prototyping and assessment of the multi-agent system.

\chedit{Another related work is the interactive LLM-based decision support system developed in \cite{lawless2024want}, which also focuses on user preferences, using meeting scheduling as a case study. In their system, the role of LLMs is to translate users' scheduling constraints into code, which is then input into a scheduling optimization package to find feasible meeting times. In contrast, our system uses LLMs to directly process members' preferences and generate suggestions. We consider our approach complementary to the system in \cite{lawless2024want}. Specifically, when constraints involve complex relationships and logical manipulations, the system in \cite{lawless2024want} can leverage optimization packages to effectively and precisely account for user preferences. However, this approach does not provide explanations for the suggestions, as optimization packages typically output only results without rationale. Moreover, optimization packages may not always be available for a given collective decision problem; our system avoids this limitation by using LLMs to directly aggregate preferences.}

\section{System Design} \label{sec:system_design}

\chedit{A group of $n$ members are to make a decision; for example, an engineering team wants to schedule a meeting with external customers to finalize their product solution.
We index the members with $i\in [n] \triangleq \{1,...n\}$.
Each member $i$ has their individual preferences about the decision.
A system powered by a large language model (LLM) is utilized to assist the group in decision-making by collecting and understanding the members' preferences and accordingly suggesting options for the group.
This coordination process proceeds at $t = 0, 1... T$ where $t=0$ is the initialization before the coordination process.
At each round, the system shows each member the current decision of the group, and each member can \emph{privately} communicate with the system about their individual preferences or opinions in their natural languages. 
}
We provide the specifics of the system and the process as follows.

\subsection{Modules}

The system consists of four components: (1) an intent extraction module, (2) a coordination module, (3) an evaluation module, and (4) \chedit{a database}.
Their functions are described below.
\begin{itemize}
      
    \item \emph{Intent Extraction Module.} \chedit{The intent extraction module is LLM-empowered and has two functions: (1) communicate with a member to solicit their opinions about the current decision of the group, (2) extract the member's individual preferences from the conversation and outputs them as a list of compact restatements.} We write $m_i^t$ for the conversation between member $i$ and the system at round $t$, which may include the system's briefings or information from previous rounds. We write $\theta_i^t$ to denote individual preferences extracted from the conversation $m_i^t$. The prompts for these two functions can be found in Appendix B.
    
    \chedit{For simplicity, we use vector notation $\vec m^t = \left ( m_1^t, \dots, m_n^t \right )$ for the collection of all the members' messages at round $t$, and similarly write $\vec \theta^t = \left ( \theta_1^t, \dots, \theta_n^t \right )$ for their individual preferences. }

    \item \emph{Coordination Module (Coordinator).} \chedit{The coordination module is LLM-empowered and receives as input the collection of the members' latest extracted individual preferences at each round $t$ (i.e., $\left ( \theta_1^t, \dots, \theta_n^t \right )$ and for simplicity, we use vector notation $\vec \theta^t$ to represent the preference profile). Based on this preference profile, the group's intent, and information from the database, the coordination module generates a list of options for the group.}  We denote the set of options at round $t$ as $\cX^t = \{ x_1^t, \dots, x_K^t \}$, where $K$ represents the number of options that the coordination module produces. 
    Furthermore, each option $x_j^t \in \cX^t$ contains three fields for informational purposes: (i) suggested decision for the group, (ii) the subset of the members for each of whom this option fulfills at least one individual preference, and (iii) a list of reasons for this suggestion.
    
     \item \emph{Evaluation Module.} \chedit{According to the preference profile $\vec \theta^t$, the evaluation module measures the quality of the options $\cX^t$ proposed by the coordination module. Specifically,
     for any option $x_j^t\in \cX^t$ and any member $i$, the module outputs a non-negative real value $\pi_{j, i}^t \ge 0$ that quantifies the degree to which the individual preferences $\theta_i^t$ of member $i$ are taken into account. We do not dictate a general framework of evaluation criteria but suggest that a feasible framework should follow two rules: (i) The value $\pi_{j, i}^t$ for option $x_j^t$ is increasing in the number of member $i$'s preferences that are fulfilled; (ii) the module assigns $\pi_{j, i}^t = 0$ if and only if none of the member $i$'s preferences in $\theta_i^t$ are fulfilled; (iii) the valuation does not depend on the identity of the members. In our system, we leverage LLMs to perform this evaluation task according to our prompt.}

     \chedit{To understand how the members' preferences are served by the system, we define three metrics that characterize the distribution of the values $\{\pi_{j, i}^t\}_{i\in[n]}$ among the members for any option $x_j^t$}. We first define the \emph{satisfaction ratio} of option $x_j^t$ as the fraction of members for whom this option meets at least one of their individual preferences, which is given by  

     \begin{align}
         r_j^t = \frac 1 n \sum_{i = 1}^n \mathbf 1 \left \{ \pi_{j, i}^t > 0 \right \},
     \end{align}
     which lies between $0$ and $1$.  
    For expositional purposes, we say that \emph{an option $x_j^t$ satisfies member $i$} when at least one of their individual preferences  ($\theta_j^t$) is fulfilled by the option, i.e., $\pi_{j, i}^t > 0$.

     We also define the \emph{satisfaction score} of the option as the average of all the members' values for the option, which measures the aggregate level to which option $x_j^t$ meets the members' preferences, as calculated by 
     \begin{align}
         s_j^t = \frac 1 n \sum_{i = 1}^n \pi_{j, i}^t.
     \end{align}

     To investigate whether the members' preferences are considered in a balanced manner, we define the \emph{equity score} of option $x_j^t$, which is given by  
     \begin{align}
         g_j^t = \begin{cases}
             1, & r_j^t = 0 \\
             \frac {\sum_{i_1 \in [n], i_2 \in [n]} |\pi_{j, i_1}^t - \pi_{j, i_2}^t|} {2n^2 \sum_{i \in [n]} \pi_{j, i}^t}, & r_j^t > 0
         \end{cases}.
     \end{align}
     The equity score is set as $1$ if option $x_j^t$ does not meet any member's preference, i.e., when $r_j^t=0$. Otherwise, the equity score is the Gini coefficient of the values $\{\pi_{j,i}^t\}_{i\in[n]}$ \cite{gini1936measure}. The equity score is between 0 and 1.

     \item \chedit{\emph{Database.} A database $\cD$ stores and provides other modules with access to the latest information about members that may be relevant for group decision-making but which members may assume as background information and therefore not explicitly provide in their conversations with the system.\footnote{\chedit{The database can be dynamically updated to reflect the latest information about the members served by the system. However, in this paper, we do not focus on the updating process. For simplicity, in our study, we assume that there is no new member information that requires consideration by the system.}} The specific content stored in the database is left to the discretion of the system users. Following our example of scheduling a meeting with external customers, the presence of both customers and engineers proposing solutions should be prioritized based on their roles and expertise. In this case, the database should include information such as members' areas of expertise, responsibilities, and their involvement in coordinating meeting times. Later in our case study, we will offer details of what our database includes. It has been shown that these heterogeneities among members have a great impact on meetings in large-scale organizations \cite{charpignon2023navigating}.}

 \end{itemize}
 
\subsection{Coordination Process}

\label{sec:coordination_process} The system assists the group in identifying and revising their decision to meet the members' preferences as much as possible through many rounds of private conversations between the members and the system.
Toward this goal, we define \emph{decision candidate} for each round $t$ as an option from the set of option $\cX^t$ that satisfies most of the group members; we denote it as $\hat x^t \in \cX^t$. 
    
Below, we describe the system protocol of how the group decision is found and improved through the $T$ rounds, as illustrated in \cref{subfig:information_flow_diagram}.
At $t=0$, we initialize  $\cX^0 = \emptyset$ and $\hat x^0 = \emptyset$. At round $t \in \{1,2...T\}$,
\begin{enumerate}
\item The system presents to each member the set of options suggested based on round $t-1$, i.e., $\cX^{t-1}$. 

\item Observing $\cX^{t-1}$, each member $i$ has their private conversation $m_i^t$ with the intent extraction module, and the intent extraction module outputs the extracted preferences $\theta_i^t$ to the coordination module. 

\item  According to the collection of members' preferences $\vec \theta^t$, the group's intent, and relevant information from database $\cD$, the coordinator generates a set of $K$ options $\cX^t$ as follows: 
\begin{itemize}
    \item For $t=1$, no candidate decision has been proposed for the group (i.e., $\hat x^{0} = \emptyset$). Therefore,  the coordinator module is instructed to suggest $K$ options, and each of the options should satisfy at least 2 members.
    
    \item  For any $t>1$, there is a decision candidate from round $t-1$ as constructed in step (4) (i.e., $\hat x^{t-1} \neq \emptyset$), the coordinator module presents $\hat x^{t - 1}$ and proposes another $K - 1$ options that should satisfy at least as many members as the candidate decision $\hat x^{t - 1}$ does.
    
\end{itemize}

\item The evaluation module generates $\pi_{j,i}^t$ for each option $x_j^t\in \cX^t$  and member $i$ based on the extracted preferences $\vec \theta^t$, and accordingly computes satisfaction ratio $r_j^t$ and satisfaction score $s_j^t$ for each option.
The decision candidate $\hat x^t$ of round $t$ is then determined by
\begin{align} \label{eq:majority}
    \hat x^t = \argmax_{x_j^t \in \cX_\star^t} s_j^t \quad \text{where} \quad \cX_\star^t = \argmax_{x_j^t \in \cX^t} r_j^t.
\end{align}
Namely, the system first identifies the set of options that satisfy the greatest number of members and then selects an option with the largest satisfaction score. We design this selection rule for the following reasons: By constructing $\cX_\star^t$, we minimize the number of members whose preferences are not fulfilled at all by the option. Then, in case there are ties between the potential decision candidates in $\cX_\star^t$, we choose the option that maximizes satisfaction score, i.e., fulfills a large number of the preferences of these members.\footnote{If multiple options still have the same maximum satisfaction score, the ties are broken arbitrarily, and one decision candidate is selected.} 
Note that the decision candidate is replaced with a new option if the new option yields a higher satisfaction ratio or satisfaction score. 

\end{enumerate}

For the remainder of the paper, we use $\hat r^t, \hat s^t, \hat g^t$ to particularly denote the satisfaction ratio, satisfaction score, and equity score of decision candidate $\hat x^t$, respectively.

\section{A Case Study: Group Meeting Scheduling} \label{sec:case_study}

We particularly study how our system performs in coordinating diverse interests, balancing trade-offs, and assisting in reaching a consensus in the context of meeting scheduling. Due to its inherent complexity and ubiquity in various social and organizational settings, meeting scheduling is a well-understood scenario for studying a collective decision-making process. (see, e.g., \cite{crawford2007experts,crawford2009learning,ephrati1994meet,benhassine2007agent,brzozowski2006grouptime,cranshaw2017calendar}). 

\begin{figure}
    \centering
    \subfigure[Meeting Invite\label{subfig:meeting_invite}]{\includegraphics[width=0.48\textwidth]{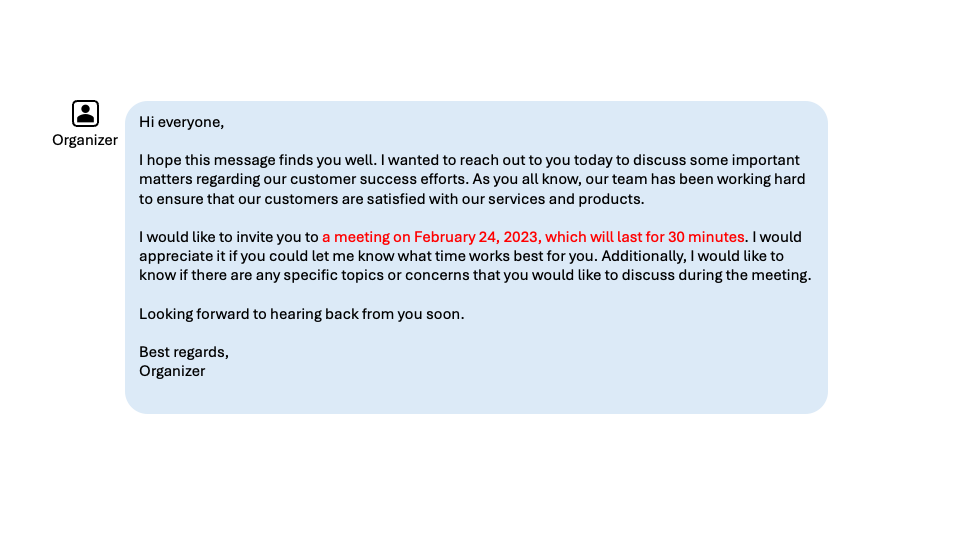}}
    \subfigure[Preference elicitation for Member 1\label{subfig:solicitor_1}]{\includegraphics[width=0.48\textwidth]{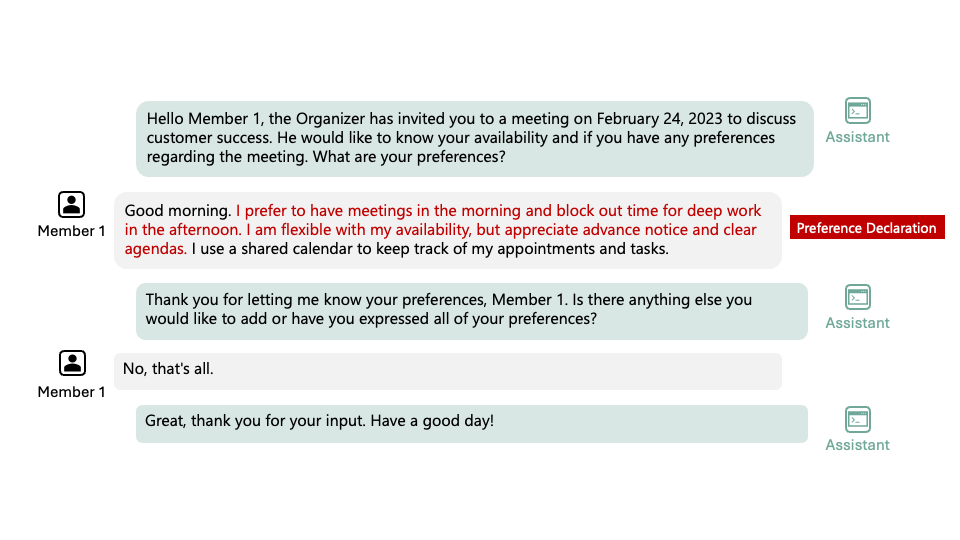}}
    \subfigure[Preference elicitation for Member 2\label{subfig:solicitor_2}]{\includegraphics[width=0.48\textwidth]{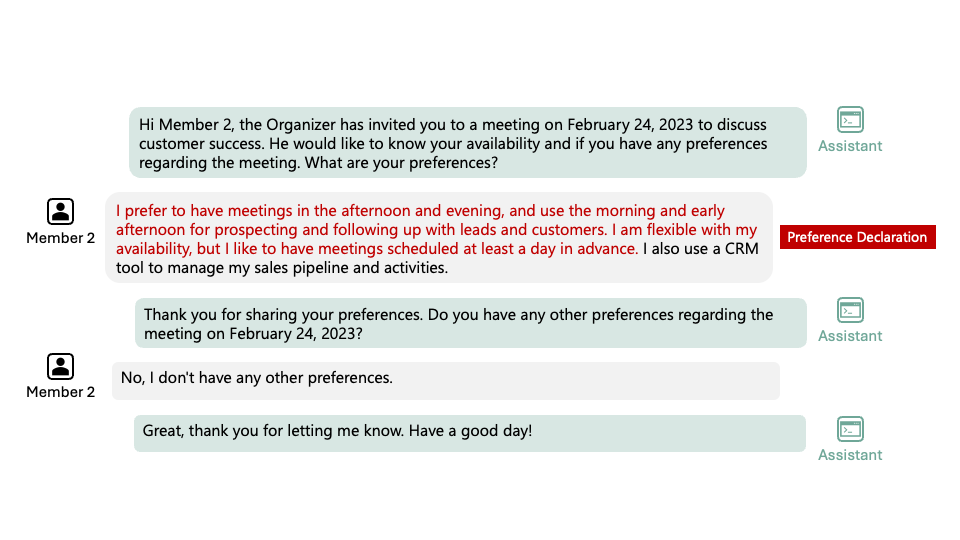}}
    \subfigure[Preference elicitation for Member 3\label{subfig:solicitor_3}]{\includegraphics[width=0.48\textwidth]{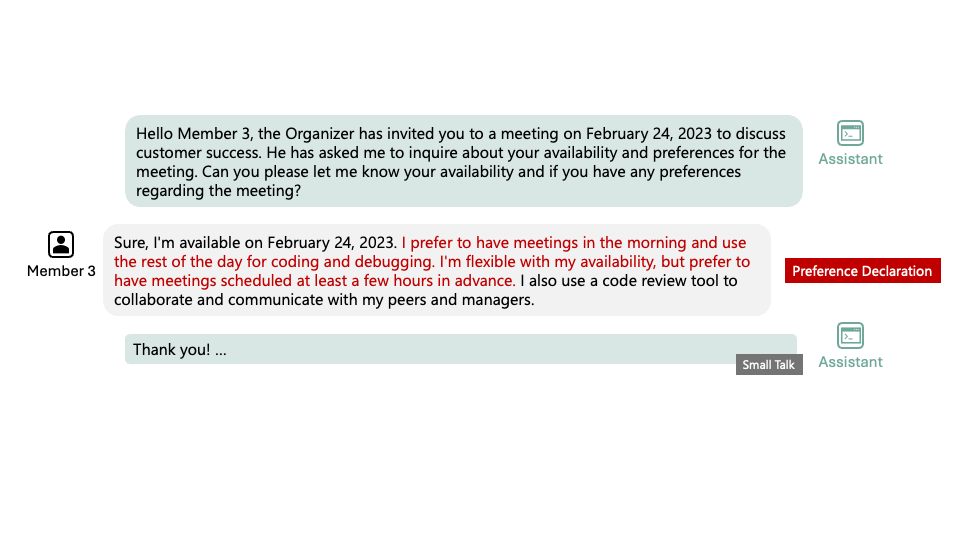}}
    \subfigure[Options proposed by the coordination module\label{subfig:coordinator}]{\includegraphics[width=0.48\textwidth]{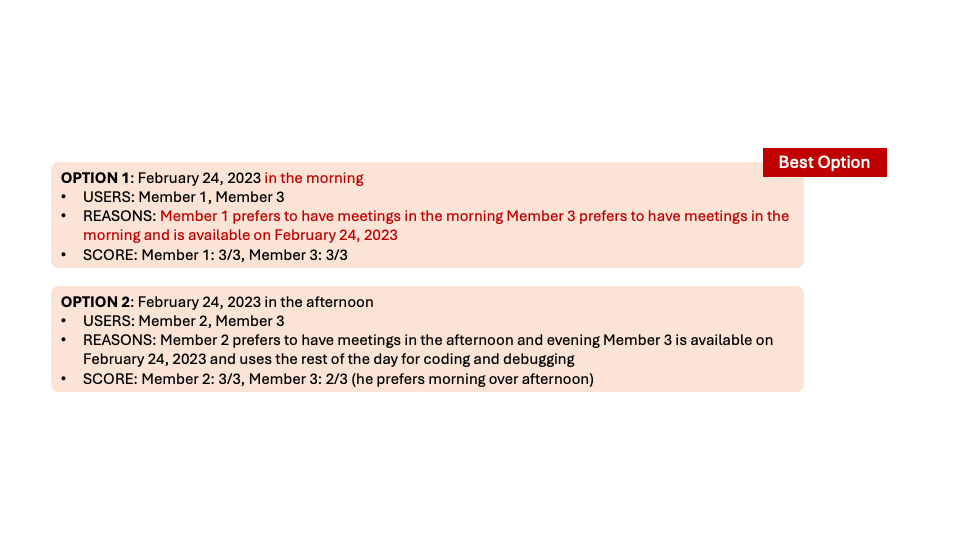}}
    \subfigure[Next round of preference elicitation and suggestion refinement for Member 1\label{subfig:solicitor_1_refinement}]{\includegraphics[width=0.48\textwidth]{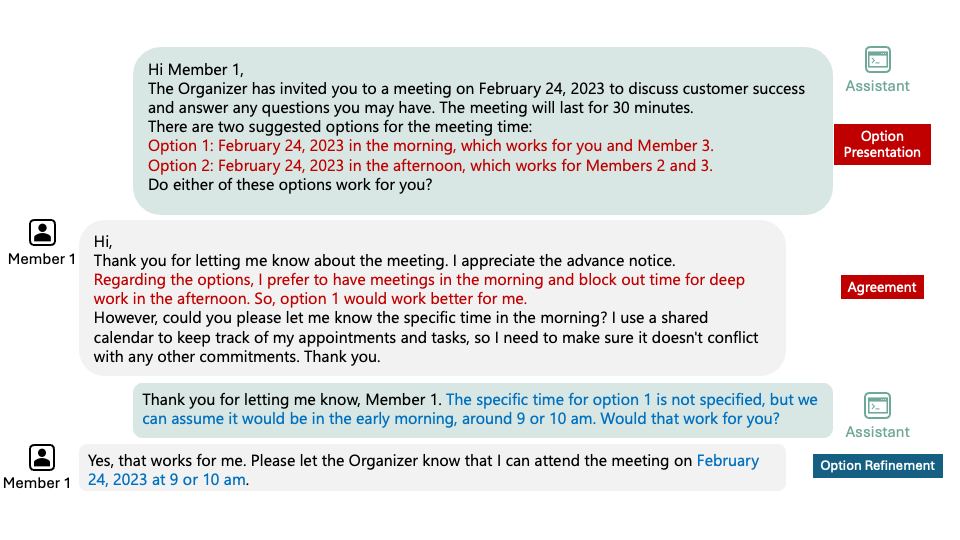}
    }    
    \caption{An example run of a coordination process for a meeting scenario.}
    \label{fig:example_scenario}
\end{figure}

We show an example scenario of meeting scheduling in \cref{fig:example_scenario}, where an organizer wants to schedule a meeting for $n = 3$ members.\footnote{In the context of meeting scheduling, we use the term ``organizer'' to refer to a member who leads the group decision process and use the ``meeting scenarios'' for the group decisions.} The meeting organizer sends out a meeting invite via the system (\cref{subfig:meeting_invite}) that describes the meeting's purpose. Next, each member converses with the intent extraction modules during which their preferences are solicited  (see \cref{subfig:solicitor_1,subfig:solicitor_2,subfig:solicitor_3}). 
The collected preferences are then sent to the coordination module for processing.

After processing the preferences, the coordination module proposes $K=2$ options as shown in \cref{subfig:coordinator} and presents them to the members.
The first option is a meeting in the morning that meets all of the revealed preferences of \chedit{Member 1 and Member 3}. The second option is an afternoon meeting, which satisfies all of Member 2's preferences and more than half of Member 3's preferences. 

The intent extraction modules start another round of conversations with the members and ask them for their opinions on the two presented options.
In \cref{subfig:solicitor_1_refinement}, Member 1 responds that the first option works better for her. Then, the module suggests refining the option by proposing a specific time for the meeting (i.e., refining ``in the morning'' to be around 9 or 10 a.m.). The system receives the members' opinions and updates their suggestions.

To implement our system for meeting scheduling leveraging LLMs, we designed a set of four prompts for the modules using \texttt{gpt-3.5-turbo}\footnote{The experiments have been run between June and August 2023.}. The details of the prompts can be found in \cref{app:prompts}. 
In particular, the evaluation module was designed to assign a value to any given member $i$ and option $x_j^t$ with Likert-scale-like values:
   \begin{align}
        \pi_{j, i}^t=
        \begin{cases}
        0, &\text{if option $x_j^t$ fulfills none of the preferences in $\theta_i^t$};\\
        1, &\text{if option $x_j^t$ fulfills at most 50\% of the preferences in $\theta_i^t$};\\
        2, &\text{if option $x_j^t$ fulfills  at least 50\% but not all of the preferences in $\theta_i^t$};\\
        3, &\text{if option $x_j^t$ fulfills  all of the preferences in $\theta_i^t$}.
        \end{cases}
    \end{align}
The satisfaction score $s_j^t$, which is derived based on $\{\pi_{j,i}^t\}_{i\in[n]}$, in turn lies between 0 and 3. 
We use this discretized scale instead of fractions (i.e., range of 0 to 1) since we did a test and empirically observed that LLMs were less capable of dealing with fractions, and errors often occurred.

We assess the performance of leveraging LLMs to coordinate meeting scheduling. We study various aspects of coordination dynamics, including the efficiency of eliciting preferences, the effectiveness of identifying mutually agreed-upon options, the satisfaction of members' preferences, and the quality of LLMs' reasoning and preference aggregation.

Specifically, we aim to answer the following research questions (RQs):  
\begin{quote}
\begin{itemize}
    \item [(RQ1)] \emph{Efficiency:} How does the length of interactions between members and the system evolve as the number of rounds progresses? Does it diminish over time?
    
    \item [(RQ2)] \emph{Effectiveness:} How many members do the options proposed by the system satisfy? Is this number increasing progressively?
    
    \item [(RQ3)] \emph{Quality:} To what extent can the system fulfill the members' preferences? Are more and more preferences gradually fulfilled? Are the preferences fulfilled in a balanced manner?
    
    \item [(RQ4)] \emph{Human-like Reasoning Ability and Preference Aggregation:} How acceptable are the reasons and the options that the system proposes? 
    
\end{itemize}
\end{quote}

We want to highlight the difference between the effectiveness and quality of the system. First, effectiveness refers to the system’s ability to involve a large fraction of members in the proposal of options, thereby minimizing the number of members whose preferences are overlooked. On the other hand, quality implies that from a set of options designed to minimize member dissatisfaction, we select the one that aligns with the greatest number of the members' preferences in a balanced manner.

A natural way to evaluate the system and answer (RQ1)-(RQ4) is by recruiting human participants and conducting a study in which the human participants use the system for meeting scheduling. However, this approach presents certain challenges in terms of implementation and scalability. To ensure the significance of the study’s results, it is necessary to conduct numerous meeting scheduling scenarios. Since each scenario involves several rounds of conversations between human participants and the system, such a study requires many human participants and time to complete. For instance, the user study of  \emph{Calendar.help} \cite{cranshaw2017calendar} lasted for 18 months. We believe that enabling a user study to be completed quickly would be beneficial in iterating over and expediting system designs.

Thus, to address these challenges, we alternatively conducted the evaluation study under synthetic meeting scheduling settings and used LLMs to simulate members' conversation behavior. This approach enables more controlled and scalable experimentation while preserving a realistic level of human-like interaction.

In \cref{sec:eval_quant}, we describe our experiments for system evaluation to answer (RQ1)-(RQ3). Specifically, we outline the simulations, the metrics, and the results. Subsequently, in \cref{sec:eval_qualitative}, we lay out the specifics of the study we conducted with human participants and present the study results to answer (RQ4).

\section{Evaluation on the Coordination Process}\label{sec:eval_quant}

\subsection{Synthetic Data}
We created a synthetic medical start-up company with 34 employees. Each employee was assigned six fields of information: (1) employee name, (2) a role (e.g., CEO, Software Engineer, Sales Manager, etc.), (3) their manager, (4) a seniority level ranging from 1 to 5, (5) responsibilities (e.g., \textit{``Assisting the sales team in various tasks and projects''}), and (6) schedule preferences (e.g., prefer to have meetings in the afternoon, to have time to drop the kids at school in the morning, etc.). \chedit{In particular, the schedule preferences are the employees' private information, but all the other information is public to the employees and the system.} \cref{tab:user_data} shows a part of example employee data.

We synthesized a collection of meetings, each of which includes: (1) an organizer, (2) members of the meetings, (3) the meeting subject, (4) date, and (5) duration. Example meeting data can be found in \cref{tab:meeting_data}. 

We constructed a social graph to represent the relationships between teammates and the collaboration among employees based on the organizational structure and the number of meetings shared between any two employees.
Specifically, teammates are a set of employees who report to the same manager.  
For each employee, we counted how often they met with every other employee. We selected the first half of the other employees who shared at least one meeting with this employee as their collaborators. \cref{tab:social_graph_meeting,subfig:meeting_network} show the social subgraph among the members\footnote{The word ``members'' corresponds to the meeting attendees.} invited to a given meeting.

\chedit{Finally, the database $\cD$ for this synthetic company encompasses all the public information about employees. The database includes the social graph and the profile data of all employees except for their schedule preferences (which are their private information and should be elicited by the system).}

In the next section, we detail our experiment setup, evaluation metrics, and results.

\subsection{Experiment Setup} \label{sec:eval_study}

We evaluated the performance of our system on the following axes: (i) the effect of the parameters of the system ($n, K$, incorporation of the database) on the performance, (ii) system performance compared to two baselines, (iii) robustness of our system subject to paraphrased preferences. 

\smallskip

\noindent \textbf{System parameters and rationale.} To evaluate our system on the former axis, we consider the following settings: 

\begin{itemize}
    \item \emph{Schedule preferences only.} First, the system was required to propose meeting options based solely on the employees’ schedule preferences \emph{without} considering the database $\cD$.
    Namely, the system focused on finding a meeting time that balances individual schedule preferences elicited from meeting members. For this setting, we examined the effect of the meeting size $n$ and the number of proposed options $K$ as follows: 

    \begin{itemize}
        \item  We first studied the effect of meeting size on the coordination processes. We set $K=2$, i.e., the coordinator could propose two options at each round. We considered meetings of three distinct sizes, specifically for $n\in\{3,4,5\}$. For each meeting size $n$, we selected 20 meeting scenarios at random from the calendar data that matched the specified size.
    
        \item Secondly, we examined the effect of the number of options $K$. We considered the meetings with size $n=5$ and considered two cases $K\in \{3,4\}$. For each number $K$, we randomly selected 20 meeting scenarios with $5$ employees for the experimentation.
    \end{itemize}

    \item \emph{Schedule preferences and database.} Second, the coordination module considered not only schedule preferences but also the information in the database $\cD$; the system also accounted for attendance constraints (if any) posed by the meeting organizer.

    To examine the effect of incorporating the database in the system's suggestions, we tested the system on 20 meeting scenarios with $n = 5$ and $K = 2$. First, we focused on $n = 5$ members. This is because, in smaller meetings ($n=3,4$), the social connections among the members were not complex enough to effectively showcase the impact of the social graph on the systems' reasoning.  Furthermore, based on the findings of the baseline study (cf. \cref{sec:results}),  the system’s performance metrics deteriorated with larger values of $K$, and, therefore, we opted for $K=2$.\footnote{\chedit{With a large $K$ the members have more options to choose, but it is also more likely that the members have diverging choices, making it difficult to refine proposed options. We will explain this observation in detail in \cref{sec:discussion}.}}
\end{itemize}

In all of the simulated meeting scenarios, the coordination processes were carried out over a total of $T = 4$ rounds. This is consistent with the finding of \emph{Calendar.help} \cite{cranshaw2017calendar}, which shows that most users had no more than five rounds of interactions with the system.

Considering the fact that most participants had meetings with 3 to 5 attendees in our requirements analysis (cf. \cref{sec:requirements_analysis}), we ran simulations with $n \in \{ 3, 4, 5 \}$ attendees. This is also consistent with the results reported in \cite{cranshaw2017calendar,yu2023large}. 

In total, we ran 120 meeting scenarios.
We note that throughout the two studies, we chose $n > K$, i.e., the meeting size was always greater than the number of proposed options. When there are more options presented at each round than the number of meeting members, it is difficult to amass a majority on one of the options and, therefore, reach a consensus.
Moreover, the coordination process is designed with the goal of collecting the members' feedback and gradually increasing the fraction of the members that the decision candidate can satisfy (i.e., satisfaction ratio $\hat r^t$). Therefore, the decision candidate is refined via several rounds of proposing new options, and proposing many options at each round is not needed. 

\smallskip

\noindent \textbf{Comparison with baselines.} Our system includes two interactive elements: (i) the conversation between a user and the intent extraction module and (ii) the multiple rounds of discussion. 
To show the gains from the two elements, we compared the system with the following two baselines:

\begin{itemize}
    \item \emph{Single-Round Non-Conversational Baseline.} In this baseline, both elements were absent. We did not use the interactive/conversational intent extraction module and instead provided preferences without processing them directly in the coordination module. We ran the protocol for only one round (i.e., $T = 1$). 

    \item \emph{Single-Round Conversational Baseline.} 
    In this baseline, we used the interactive intent extraction module for the conversation with users but ran the system for only one round. This is to measure the extent to which multiple rounds of refining the decision candidate improves the system performance.
\end{itemize}

\smallskip

\noindent \textbf{Robustness to paraphrased preferences.} We performed a robustness check where we paraphrased the schedule preferences in the simulations using the LLM and observed how the metrics would be affected. Specifically, we considered the scenarios with $n = 3$ members and $K = 2$ options after zero, one, and two paraphrases of the preferences. To measure the metrics of the system, we considered the Single-Round Non-Conversational baseline since it did not contain the intent extraction module, which might otherwise affect the evaluation.

\subsection{Evaluation Metrics} We derived the following metrics for each round of  coordination processes to gauge the performance of our system:
\begin{itemize}
    \item \emph{Average number of interactions of a member with the system.} This measures the average number of turns in the conversation between a member and the intent extraction module. This metric is used to assess the efficiency of our system (cf. RQ1). In an efficient system, we expect the average number of interactions to decrease in the number of rounds.   
    \item \emph{Satisfaction ratio of the decision candidate ($\hat r^t$).} This metric measures the fraction of the meeting members the decision candidate satisfies. In an effective system (cf. RQ2), this ratio is expected to increase in the number of rounds, i.e., the decision candidate satisfies more members, and the system is improving its suggestions.
 
    \item \emph{Satisfaction score of the decision candidate ($\hat s^t$).} This metric quantifies the average level that the decision candidate meets a member's preferences. An increasing trend of this metric suggests the high quality of suggested options (cf. RQ3) since the system can identify options that meet many of the members' preferences. 
    
    \item \emph{Equity score of the decision candidate ($\hat g^t$).} The equity score reflects the extent to which the system balances members' preferences. We expect that in a system of good quality (cf. RQ3), the equity score of the decision candidate should decrease at every round. This means that the scores assigned to each member by the evaluation module have gradually smaller statistical dispersion as the coordination progresses, i.e., the members' preferences are accommodated in a more balanced way.
    
\end{itemize}

\section{Simulation Study Results} \label{sec:results}

\begin{figure}[t]
    \centering
    \includegraphics[height=5cm,keepaspectratio]{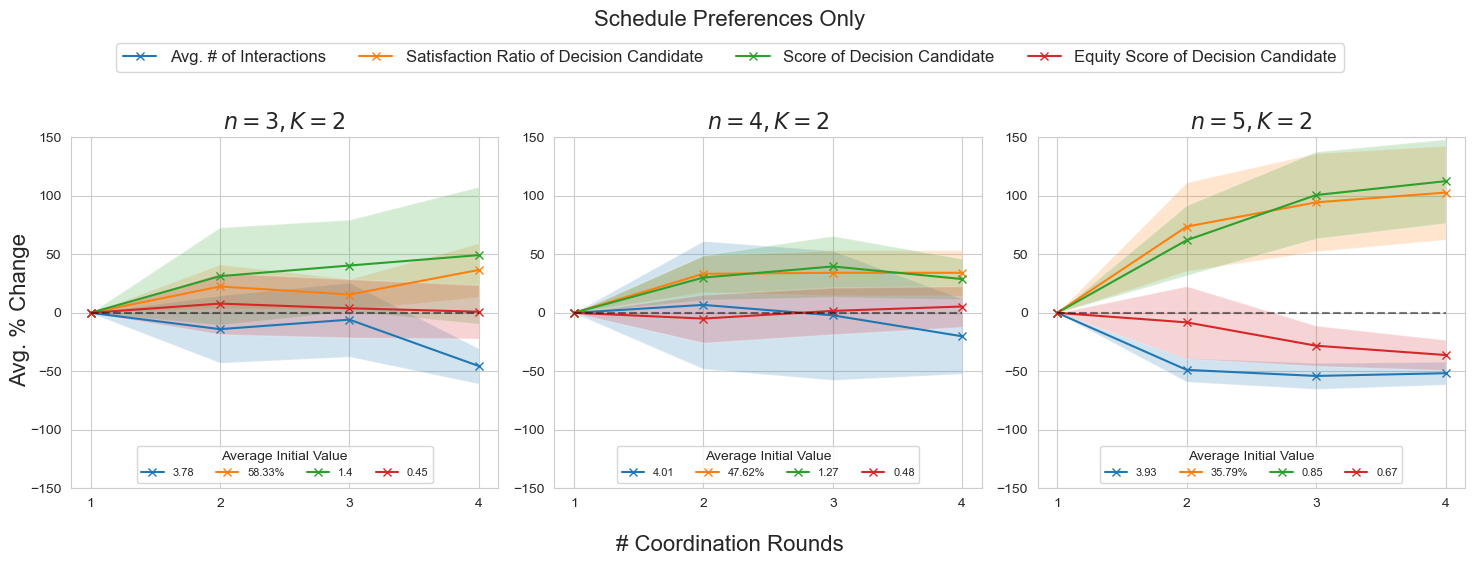}
    \caption{We plot the average of the percentage change in the metrics with the number of rounds for the settings $T=4, K=2$ and $n\in\{3,4,5\}$. For each setting, 20 different meeting scenarios were used. Given the metrics of a meeting scenario, the percentage change for each round is computed with respect to $t = 1$. The plots show the average of the percentage changes over the 20 scenarios.  The legends in the \chedit{bottom side} of each sub-figure correspond to the \chedit{average value of the metrics at $t=1$. The percentage changes of all the metrics when $t=1$ are all zero, given there are no references from the previous round.}}
    \label{fig:trend_lines_results_homogeneous}
\end{figure}

\begin{figure}[t]
    \centering
    \includegraphics[height=5cm,keepaspectratio]{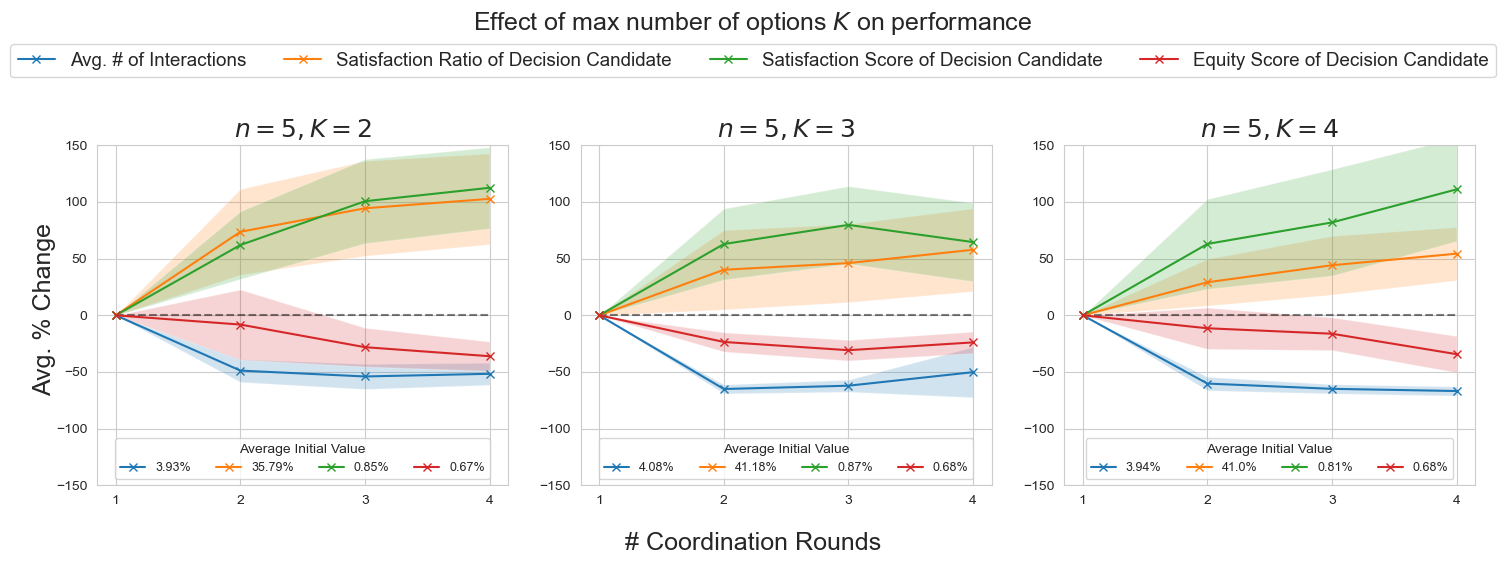}
    \caption{The average of the percentage change in each metric with the number of rounds for $T=4,n=5$ and $K\in\{2,3,4\}$. For each setting, 20 different meeting scenarios were tested.}
    \label{fig:trend_lines_results_homogeneous_options}
\end{figure}

In this section, we present the results of our study in \cref{sec:eval_quant}. \cref{fig:trend_lines_results_heterogeneous,fig:trend_lines_results_homogeneous,fig:trend_lines_results_homogeneous_options} summarize the evaluation metrics in the experiment of \cref{sec:eval_quant}, which are plotted in the following manner: Given a meeting scenario, for each metric we calculated its percentage change at round $t$ with respect to its value at round $t=1$. Then we plotted the average of the percentage change in the metrics for each round, together with their 95\% confidence intervals in our system\footnote{We use the Student-t distribution with 19 degrees of freedom and a two-sided confidence interval.}. The plots showing the metrics values for all scenarios are relegated to \cref{app:coordinator_results}.

\subsection{Effect of Meeting Size $n$} \cref{fig:trend_lines_results_homogeneous} shows the results when meeting size $n$ varies. We observed three trends across the three settings examined:
 (i) the average number of \chedit{interactions} is decreasing, (ii) the satisfaction ratio is increasing, (iii) the satisfaction score is increasing. 
 
In particular, the system performed significantly better in these three dimensions in the setting with $n = 5$ members than in the ones with $n = 3$ and $n = 4$.
For $n = 5$ members, the average number of interactions was largely reduced by $\approx 50\%$\footnote{We use ``$\approx$'' to for ``approximately''.}, compared to $\approx 25\%$ ($n = 3$) and $\approx 45\%$ ($n = 4$).
Secondly, the satisfaction score of the decision candidate for $n=5$ was eventually increased by $\approx 120\%$, while it was increased by
 $\approx 45\%$ ($n = 3$) and  $\approx 40\%$ ($n = 4$) respectively. Similarly, the satisfaction ratio of the decision candidate was increased by $\approx 110\%$ for $n = 5$, while both $n=3$ and $n=4$ experienced an increase of $\approx 40\%$. 
Finally, the equity score of the decision candidate decreased for $n = 5$, but it exhibited subtle fluctuations for $n = 3$ and $n = 4$. 

\subsection{Effect of Number of Options $K$} 
\cref{fig:trend_lines_results_homogeneous_options} shows the results of the experiment for the settings when the number of options $K$ varies. 
We found that increasing the number of options from $K=2$ to $K=4$ decreased the final percentage change of the satisfaction ratio.
Specifically, for $K = 2$ options, the system achieved an $\approx 100\%$ increase in the satisfaction ratio at the final round, whereas adopting $K = 3$ and $K = 4$ options yielded an increase of $\approx 50\%$.      
We also found that the increase in satisfaction score was $\approx 100\%$ for $K=2$, compared to $\approx 50\%$ ($K=3$) and $\approx 100\%$ ($K = 4$). The equity score experienced a smaller percentage decrease as $K$ was increased. These observations indicated that the system's performance was negatively impacted as the number of options $K$ was increased. 

\begin{figure}[t]
    \centering
    \includegraphics[height=5cm,keepaspectratio]{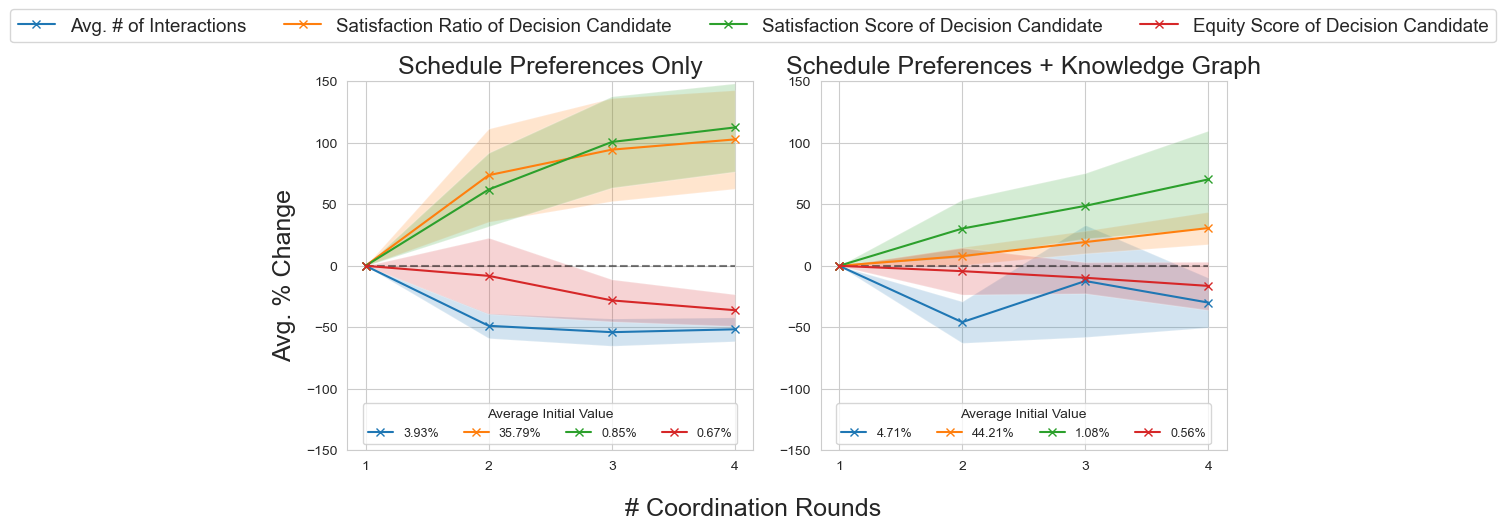}
    \caption{The average of the percentage change in each metric with the number of rounds for $T=4,n=5, K=2$ and whether the database $\cD$ is considered or not. For each setting, 20 different meeting scenarios were tested. }
    \label{fig:trend_lines_results_heterogeneous}
\end{figure}

\subsection{Effect of Incorporating the database} 
\cref{fig:trend_lines_results_heterogeneous} compares the metrics between including and excluding the database for $T=4$, $n=5$ and $K=2$. 
When the database and attendance constraints were incorporated, the average number of interactions remained decreasing, and the satisfaction ratio and the decision candidate's score increased. However, compared to using the schedule preferences only, the evaluation metrics deteriorated, and the system's performance worsened in each dimension.

Precisely, the decrease in the average number of interactions was reduced from $\approx 50\%$  to an $\approx 25\%$.  The final percentage increase in the satisfaction ratio of the decision candidate dropped from $\approx 110\%$ to less than $\approx 50\%$. Moreover, the final percentage increase in the satisfaction score decreased from  $\approx 120\%$ to $\approx 75\%$. 
    
This pattern is expected since introducing multiple additional constraints embedded in the database's complexity creates more challenges in balancing the members' preferences. For example, suppose that one of the members must attend a meeting. Any options must accommodate this requirement, which may, in turn, yield options that are not satisfactory or require compromises (i.e., still feasible but with fewer preferences satisfied) for the other members.

\begin{table}[t]
    \centering
    \tiny
    \caption{Numerical results from comparison with baselines. The averages are reported with a 95\% confidence interval. \textuparrow\;(resp. \textdownarrow) means the higher (resp. the lower), the better.}
    \begin{tabular}{lrrr}
    \toprule
        ~ & Single-Round Non-Conversational  & Single-Round Conversational & \textbf{Our Proposed System} \\ 
    \midrule
        ~ & \multicolumn{3}{c}{$n = 3, K = 2$} \\ 
    \midrule
        Satisfaction Ratio (\%, \textuparrow) & 57.14 $\pm$ 6.86 & 58.33 $\pm$ 5.58 & \bf{78.33 $\pm$ 6.15} \\ 
        Satisfaction Score (\textuparrow) & 1.7 $\pm$ 0.17 & 1.4 $\pm$ 0.17 & \bf{1.73 $\pm$ 0.2} \\ 
        Equity Score (\textdownarrow) & 0.41 $\pm$ 0.07 & 0.45 $\pm$ 0.06 & \bf{0.35 $\pm$ 0.06} \\ 
        Avg. Number of Interactions (\textdownarrow) & n/a & 3.78 $\pm$ 0.8 & \bf{2.35 $\pm$ 1.11} \\
    \midrule
        ~ & \multicolumn{3}{c}{$n = 4, K = 2$} \\ 
    \midrule
        Satisfaction Ratio (\%, \textuparrow) & 58.33 $\pm$ 8.37 & 47.62 $\pm$ 4.01 & \bf{63.1 $\pm$ 6.24} \\ 
        Satisfaction Score (\textuparrow) & \bf{1.81 $\pm$ 0.17} & 1.27 $\pm$ 0.16 & 1.5 $\pm$ 0.12 \\ 
        Equity Score (\textdownarrow) & \bf{0.35 $\pm$ 0.07}  & 0.48 $\pm$ 0.06 & 0.43 $\pm$ 0.05 \\ 
        Avg. Number of Interactions (\textdownarrow) & n/a & 4.01 $\pm$ 0.97 & \bf{2.18 $\pm$ 0.54} \\ 
    \midrule
        ~ & \multicolumn{3}{c}{$n = 5, K = 2$} \\ 
    \midrule
        Satisfaction Ratio (\%, \textuparrow) & 48.57 $\pm$ 4.39 & 35.79 $\pm$ 4.15 & \bf{65.26 $\pm$ 6.75} \\ 
        Satisfaction Score (\textuparrow) & 1.43 $\pm$ 0.15 & 0.85 $\pm$ 0.13 & \bf{1.64 $\pm$ 0.18} \\ 
        Equity Score (\textdownarrow) & 0.5 $\pm$ 0.06 & 0.67 $\pm$ 0.07 & \bf{0.36 $\pm$ 0.05} \\ 
        Avg. Number of Interactions (\textdownarrow) & n/a & 3.93 $\pm$ 0.53 & \bf{1.75 $\pm$ 0.34} \\
    \midrule
        ~ & \multicolumn{3}{c}{$n = 5, K = 3$} \\ 
    \midrule
        Satisfaction Ratio (\%, \textuparrow) & 46.00 $\pm$ 4.95 & 41.18 $\pm$ 5.41 & \bf{56.47 $\pm$ 5.22} \\ 
        Satisfaction Score (\textuparrow) & \bf{1.30 $\pm$ 0.17} & 0.87 $\pm$ 0.15 & 1.26 $\pm$ 0.17 \\ 
        Equity Score (\textdownarrow) & 0.53 $\pm$ 0.06 & 0.68 $\pm$ 0.05 & \bf{0.51 $\pm$ 0.06} \\ 
        Avg. Number of Interactions (\textdownarrow) & n/a & 4.08 $\pm$ 0.36 & \bf{1.82 $\pm$ 0.55} \\
    \midrule
        ~ & \multicolumn{3}{c}{$n = 5, K = 4$} \\ 
    \midrule
        Satisfaction Ratio (\%, \textuparrow) & 52.0 $\pm$ 7.89 & 41.0 $\pm$ 5.17 & \bf{60.00 $\pm$ 6.47} \\ 
        Satisfaction Score (\textuparrow) & 1.38 $\pm$ 0.19 & 0.81 $\pm$ 0.13 & \bf{1.41 $\pm$ 0.16} \\ 
        Equity Score (\textdownarrow) & 0.52 $\pm$ 0.07 & 0.68 $\pm$ 0.06 & \bf{0.39 $\pm$ 0.07} \\ 
        Avg. Number of Interactions (\textdownarrow) & n/a & 3.94 $\pm$ 0.41 & \bf{1.2 $\pm$ 0.11} \\
    \midrule
        ~ & \multicolumn{3}{c}{$n = 5, K = 4$ (w/ database)} \\ 
    \midrule
        Satisfaction Ratio (\%, \textuparrow) & 46.67 $\pm$ 4.84 & 44.21 $\pm$ 3.24 & \bf{56.84 $\pm$ 5.33} \\ 
        Satisfaction Score (\textuparrow) & 1.43 $\pm$ 0.18 & 1.08 $\pm$ 0.16 & \bf{1.57 $\pm$ 0.17} \\ 
        Equity Score (\textdownarrow) & 0.43 $\pm$ 0.07 & 0.56 $\pm$ 0.06 & \bf{0.41 $\pm$ 0.06} \\ 
        Average Number of Interactions (\textdownarrow) & n/a & 4.71 $\pm$ 1.23 & \bf{2.43 $\pm$ 0.60} \\ 
    \bottomrule
    \end{tabular}
    \label{tab:baselines}

\end{table}

\subsection{Comparison with Baselines and Robustness Checks}

\cref{tab:baselines} reports the performance of the baselines and our proposed method. First, we observed that our proposed system outperformed the two baselines in almost every metric, suggesting that the conversational multi-round decision process yielded better decision candidates across all metrics. Second, we found that having the intent extraction module had a negative impact on the performance of the system in the first round; however, repeating the process for another several rounds yielded considerable improvements across all metrics. 

The results are as expected since, by construction, our system aims to improve the satisfaction ratio iteratively. We also found that the improvement in the satisfaction ratio was correlated with improvements in the other metrics (i.e., satisfaction score, equity score, and the average number of interactions).

Finally, in \cref{tab:robustness} we found that the results remain almost unchanged after repeatedly paraphrasing the schedule preferences.
This suggested our system was robust to the paraphrased preferences.

\begin{table}[]
    \centering
    \caption{Effect of paraphrased schedule preferences for the scenario with $n = 3$ and $K = 2$.}
    \scriptsize
    \begin{tabular}{lrrr|r}
        \toprule
        Number of Perturbations & 0 & 1 & 2 & Std. of the Mean \\
        \midrule
        Satisfaction Ratio & 57.14 $\pm $ 6.86 & 58.73 $\pm$ 6.60 & 60.32 $\pm$ 6.27 & 1.25 \\
        Satisfaction Score & 1.70 $\pm$ 0.17 & 1.65 $\pm$ 0.20 & 1.67 $\pm$ 0.19 & 0.025 \\
        Equity Score & 0.41 $\pm$ 0.07 & 0.41 $\pm$ 0.07 & 0.40 $\pm$ 0.07 & 0.005 \\ 
        \bottomrule
    \end{tabular}

    \label{tab:robustness}
\end{table}

\section{Evaluation on Reasoning and Option Suggestion}\label{sec:eval_qualitative}

For the employees who signed up for our study (see \cref{sec:requirements_analysis}), we randomly assigned them to three groups (A, B, C) and sent each of them the online survey corresponding to their group via email. 
We collected a total of 45 effective responses: Group A, 14 responses; Group B, 13 responses; Group C, 18 responses. All the detailed information of this study can be found in \cref{app:user_study}.

\subsection{Procedure and Tasks} 

We initially divided the employees who had expressed their interests into three distinct groups (A, B, C). For each group, we distributed a survey that began with eligibility questions followed by a consent form for their participation in our study. Upon receiving informed consent from the eligible participants, they proceeded to complete the main questionnaires, which were estimated to take between 25 and 35 minutes. 
The content of the questionnaires was tailored according to the specific group to which the participants were assigned, as described below.

All the questionnaires were designed based on a single synthetic meeting scenario involving $n = 3$ members from our database, along with $K=2$ proposed options generated after the first round of the coordination process.
At the outset of the questionnaires, the participants were presented with the meeting invite, the two proposed options the members each option satisfies, and the corresponding reasons. Participants were then tasked with assessing the results in this meeting scenario in three parts, from the perspective of one of the meeting members: Group A evaluated them as the first member, Group B as the second, and Group C as the third.

\begin{enumerate}

    \item In the first part of the survey, the participants were provided with only the preferences of \emph{the member corresponding to their group}, and then they were asked to evaluate the options proposed by the system from the perspective of their corresponding member. 

    \item In the second part, the participants were instead given with \emph{all the members' preferences} and were required to answer the same set of evaluation questions from the perspective of their corresponding member.

    \item In the third part, the participants were asked to share their impression of the coordination system. Specifically, the participants were asked to choose how many options the system should propose for meetings of size $n = 3$ and $n = 5$ from their own perspective and provide the level of their confidence in using such a system for meeting scheduling purposes. 
\end{enumerate}


\subsection{Measures and Analysis}

In the survey questions, we used a 5-point Likert scale ranging from 1 to 5, with 1 representing the poorest performance and 5 representing the best rating. We also used free-text questions and asked the participants for their free-text feedback on their evaluation.

For each question that elicits numerical ratings, we computed the average of the responses of the participants in two ways: We calculated the average for all participants and the average for each group. 
For each of the averages, we calculated a 95\% confidence interval (two-sided) in our system using the Student-t distribution with 44 degrees of freedom for all participants, 13 degrees of freedom for Group A, 12 degrees of freedom for Group B, and 17 degrees of freedom for Group C.

\begin{table}[t]
    \centering
    \caption{Aggregate results of our survey study on reasoning and option suggestion. The average rating is reported with a 95\% confidence interval. The detailed scenario and questions are in \cref{app:user_study}.}
    
    \scriptsize
\begin{tabular}{p{0.7\textwidth}r}
\toprule
 Question                                                 & Average Rating   \\
\midrule
\multicolumn{2}{c}{All groups (45 responses)} \\
\midrule
To what degree do you believe that the system takes your preferences into account when presenting options?                                                         & 3.51 $\pm$ 0.35   \\
 Do you find the reasons behind the options acceptable based on your preferences?                                                                                                  & 3.32 $\pm$ 0.37   \\
 After seeing all members' individual preferences, to what degree do you believe that our system takes your preferences into account when presenting options?      & 3.54 $\pm$ 0.35   \\
 After seeing all members' individual preferences, do you find the reasons behind the options acceptable?                                                                        & 3.46 $\pm$ 0.36   \\
 How well do you agree the options presented reflect the preferences of the members?                                                                                             & 3.07 $\pm$ 0.32   \\
 How confident are you in using our system to coordinate members' preferences in meeting scheduling?                                                                    & 3.49 $\pm$ 0.33   \\

\midrule
\multicolumn{2}{c}{Group A (14 responses)} \\
\midrule

To what degree do you believe that the system takes your preferences into account when presenting options?                                                    & 3.64 $\pm$ 0.59   \\
 Do you find the reasons behind the options acceptable based on your preferences?                                                                                             & 3.36 $\pm$ 0.59   \\
 After seeing all members' individual preferences, to what degree do you believe that our system takes your preferences into account when presenting options? & 3.91 $\pm$ 0.60   \\
 After seeing all members' individual preferences, do you find the reasons behind the options acceptable?                                                                   & 4.09 $\pm$ 0.53   \\
 How well do you agree the options presented reflect the preferences of the members?                                                                                        & 3.36 $\pm$ 0.66   \\
 How confident are you in using our system to coordinate members' preferences in meeting scheduling?                                                               & 4.09 $\pm$ 0.53   \\
 
\midrule
\multicolumn{2}{c}{Group B (13 responses)} \\
\midrule
 To what degree do you believe that the system takes your preferences into account when presenting options?                                                    & 3.17 $\pm$ 0.73   \\
 Do you find the reasons behind the options acceptable based on your preferences?                                                                                             & 3.08 $\pm$ 0.80   \\
 After seeing all members' individual preferences, to what degree do you believe that our system takes your preferences into account when presenting options? & 3.67 $\pm$ 0.70   \\
 After seeing all members' individual preferences, do you find the reasons behind the options acceptable?                                                                   & 3.50 $\pm$ 0.76   \\
 How well do you agree the options presented reflect the preferences of the members?                                                                                        & 3.08 $\pm$ 0.61   \\
 How confident are you in using our system to coordinate members' preferences in meeting scheduling?                                                               & 3.33 $\pm$ 0.75   \\
 
 \midrule
\multicolumn{2}{c}{Group C (18 responses)} \\
\midrule
  To what degree do you believe that the system takes your preferences into account when presenting options?                                                          & 3.67 $\pm$ 0.60   \\
 Do you find the reasons behind the options acceptable based on your preferences?                                                                                                   & 3.44 $\pm$ 0.65   \\
 After seeing all members' individual preferences, to what degree do you believe that our system takes your preferences into account when presenting options?       & 3.22 $\pm$ 0.59   \\
 After seeing all members' individual preferences, do you find the reasons behind the options acceptable?                                                                         & 3.06 $\pm$ 0.59   \\
 How well do you agree the options presented reflect the preferences of the members?                                                                                              & 2.89 $\pm$ 0.55   \\
 How confident are you in using our system to coordinate members' preferences in meeting scheduling?                                                                     & 3.22 $\pm$ 0.51   \\

\bottomrule
\end{tabular}

    \label{tab:user_study}
\end{table}

\subsection{Evaluation Results} \label{sec:qualitative_results}
\cref{tab:user_study} presents the average rating, along with a 95\% confidence interval, that we received from human participants in response to a variety of questions for a specific meeting scenario (refer to Appendix D). The ratings are provided in aggregate (i.e., all groups combined) and also broken down by participant group (A, B, C).

\begin{itemize}
    \item \emph{Perceived system performance in reasoning.} Overall, the participants evaluated the system's performance as above average (i.e., which corresponds to a Likert scale value $\ge 3$), indicating that the participants liked the system.
 Most of the participants expressed satisfaction with our system, appreciating its ability to provide acceptable reasons. However,  they also pointed out that some options might be more suitable for a larger number of members than those suggested by the system. Additionally, they noted that a few preferences may carry more weight than others.
 We also observed that the participants had different interpretations of the same schedule preferences and reasons for the proposed options, leading to disagreements on whether a particular option satisfied a member’s needs. 

    \item \emph{Confidence in using our system.} The participants rated their confidence in using our system for meeting scheduling as above average, ranging from an average rating of 3.22 (for Group C) up to a rating of 4.09 (for Group A).
    
    \item \emph{Preferred number of suggested options.}  
    When asked about the number of suitable options for meetings with $n=3$ and $n=5$ members, the majority of participants indicated that $K=3$ options would be the best in both scenarios (57\% of the participants for $n = 3$, and 64\% for $n = 5$). This opinion was expressed despite the survey presenting only $K=2$ options, which our synthetic simulations suggested as the best choice since increasing the number of options diminishes the system’s performance (cf. \cref{fig:trend_lines_results_homogeneous_options}).
    We believe that this is understandable because humans may want more available options to choose from, but they might not be aware that an increase in options could potentially complicate the process of reaching a consensus.

    \item \emph{Additional feedback from human participants.} As future extensions of the system, the participants offered several suggestions on how to improve our system (overall and specific to meeting scheduling), such as the incorporation of negative reasons (i.e., potential blocks and conflicts and their severity), handling information leakage (i.e., avoiding exposing sensitive preferences to the other members), the priorities of preferences, learning from the members' previous behavior. In particular, for meeting scheduling, the participants mentioned the incorporation of timezones, availability, working style, and work plan (i.e., remote, in-person, hybrid).

\end{itemize}



\section{Discussion} \label{sec:discussion}

\subsection{Interpreting the Results} \label{sec:resultinterpretation}

Our evaluation results demonstrated that our system excels across a wide range of metrics. Specifically, it has performed well in addressing all the research questions posed.

First, in terms of efficiency (RQ1), we observed that the average number of interactions showed a substantial reduction of approximately 50\% in most settings (cf. \cref{fig:trend_lines_results_heterogeneous,fig:trend_lines_results_homogeneous,fig:trend_lines_results_homogeneous_options}). 
This reduction signified the efficiency of our system, highlighting a decrease in the number of interactions between the members and the intent extraction module.
This resulted from our system’s ability to consolidate preferences and suggested options: The members found the proposed options to be satisfactory and did not need to refine the options, leading to a reduced number of interactions in later rounds.

Second, in terms of effectiveness (RQ2), most of the experiments showed consistent progression in the satisfaction ratio of the decision candidate (cf. \cref{fig:trend_lines_results_heterogeneous,fig:trend_lines_results_homogeneous,fig:trend_lines_results_homogeneous_options}). This outcome demonstrated that the coordination module was successful in its task of progressively refining its proposed options to satisfy more members. The effectiveness of the system was also exemplified in scenarios where a meeting involved many members and finding a mutually agreeable meeting time was challenging (cf. $n=5$ in \cref{fig:trend_lines_results_homogeneous_options}).
We noted that adopting a smaller number of options $K$ achieved higher satisfaction ratios, as evidenced in \cref{fig:trend_lines_results_homogeneous_options}.
Lastly, when attendance constraints and the database were integrated (cf. \cref{fig:trend_lines_results_heterogeneous}), the system continued to improve the satisfaction ratio, albeit at a significantly reduced magnitude. This suggested that despite the added heterogeneities and complexity, the system was still able to accommodate members’ preferences, though its performance was less effective.

Third, regarding the quality of proposed options (RQ3), our findings showed that the decision candidate's satisfaction scores improved across the settings.
In particular, we observed that as the number of members increased (cf. \cref{fig:trend_lines_results_homogeneous}), there was a large improvement in the percentage change in the satisfaction score. 
On the other hand, when the number of options was varied from $K = 2$ to $K = 3$ or $4$, the satisfaction scores were reduced (cf. \cref{fig:trend_lines_results_homogeneous_options}). 
It was inconclusive whether the options that were successively proposed by the coordination had lower equity scores. 

Below, we explain the subtle effect of the number of options $K$ on the system's performance in terms of satisfaction ratio and score. 
In \cref{fig:trend_lines_results_homogeneous_options}, it was observed that an increase in the value of $K$ corresponded to a higher initial satisfaction ratio and score (at $t = 1$). This is because when there are more options to propose, it becomes easier to suggest an option that aligns with the preferences of a larger number of members.
Nonetheless, a larger $K$ impedes the system process of refining the proposed options at later rounds.
After the initial round, the members may explicitly express their preferences over the options proposed at the previous rounds, especially when several options simultaneously satisfy their needs.\footnote{In our example, Member 1 specifically favored option 1 as shown in  \cref{subfig:solicitor_1_refinement}.}
When there are more options (i.e., a larger $K$), it is more likely that the members favor different options. This, in turn, can pose a challenge for the coordination module as it attempts to refine the options (i.e., propose new options to improve the ratio and score) while accommodating these varied preferences. 
Consequently, the potential improvement in the satisfaction ratio and score may be diminished.
\footnote{We offer an example to illustrate this degradation in the system's performance when $K$ is larger. Consider a meeting with $n = 3$ members (1,2,3) and $K = 3$. Three options (say, A, B, and C) are proposed in the first round, where option A satisfies members 1 and 2, B satisfies members 2 and 3, and C satisfies members 3 and 1. In the next round, the three members can tell the system what they favor among options A to C. It can happen that each member favors a different option from the others (e.g., A favors 1, B favors 2, and C favors 3), creating a challenge for the coordinator to refine which option since the distribution of the members' preference over the options is uniform.
In contrast, if $K$ was decreased, say to $2$, such uniform distribution is less likely to occur because one of the two options tends to be preferred by at least two members more than the other. Therefore, the coordinator can focus on refining this option and progressively improving the satisfaction ratio and score.}

\chedit{The reasoning similarly explains why there were better improvements in the satisfaction ratio and score when the size of the meeting $n$ was increased (cf. \cref{fig:trend_lines_results_homogeneous}).}
For small $n$, it is likely that several options can satisfy the largest number of members, and the system needs another few rounds to decide which option to refine. We use the example in \cref{fig:example_scenario} for illustration. 
In the first round, the proposed option A satisfies Member 1 and Member 3, and option B satisfies Member 2 and Member 3; namely, both options satisfy 2 members. Suppose that the evaluator selects option A as the decision candidate. At the next round, the members express their opinions about the options. If Member 3 clearly prefers option A, then option A is indeed more favored by the group members. Consequently, the system can proceed to refine this candidate by suggesting a new option. However, if Member 3 prefers option B, the system recognizes that the members are more inclined towards option B, and it will propose option B again (along with the candidate option A). At the end of this round, option B will supersede A as the decision candidate for subsequent rounds. This impedes progress during the coordination process as the system seeks to resolve ties by gathering additional opinions. 
In contrast, with larger meeting sizes, the instances where ties need further resolution tend to occur less frequently, resulting in less friction during the process.

Lastly, the participants rated the system as having better-than-average reasoning capabilities and producing options with clear explanations. However, due to their subjective interpretation of the members’ preferences, there were some disagreements about the suitability of certain options to meet the members’ needs. This suggested that the LLM-based system might propose options following processes akin to human reasoning, which mostly focuses on finding good feasible solutions, in contrast to the optimization modules a scheduling client would use. We are confident that LLMs have the potential to suggest interpretable and feasible solutions for a wide range of optimization problems, extending beyond just meeting scheduling.\footnote{See \cite{yang2023large} for their use of LLMs for optimization problems.}

\subsection{Ablation Study}

A natural question arises about our system design: \emph{How do different components of the system affect the system performance?} We derived several insights from our experiments.

Firstly, as shown in \cref{tab:baselines}, the intent extraction module is essential to the functionality of the system both in terms of user-friendliness as well as system performance. Specifically, we observed that even though with the intent extraction module, the performance was worse in the first round (cf. the performance comparison between the two baselines in \cref{tab:baselines}), the effect of this module became positive when the coordination proceeds in rounds. \chedit{In the initial round, communication between members and the system may be unclear, potentially leading to suboptimal results. However, as coordination progresses, the intent extraction module can better understand members' preferences by asking clarifying questions. In later rounds, members can also propose adjustments to existing options through the intent extraction module, thereby improving system metrics.} 
Moreover, we found that the system was robust to the paraphrased preferences, indicating that the intent extraction module effectively extracted the same intents from similarly phrased preferences.

Secondly, allowing for several rounds of coordination resulted in better performance than a single-round process across all metrics (cf. \cref{tab:baselines}), since the coordination module could improve the decision candidate by soliciting more opinions and refining existing options. 

Thirdly, human participants' positive assessments of the system aligned with the scores assigned by the evaluation module, showcasing the effectiveness of the evaluation module in measuring satisfaction.

Finally, including the database could capture more complexities and heterogeneities (e.g., social dynamics, responsibilities, etc.) for the decision-making process; however, this also had negative impacts on the performance metrics, as discussed in \cref{sec:results}.

\subsection{Challenges and Limitations in System Implementation}
We point out several challenges of implementing this system in practical use for work coordination and collective decision-making.

The first set of challenges involves token limits, the high number of requests the system sends to the LLMs, and the latency of processing the language input when the meeting size is large. A potential solution could be calibrating the coordination protocol. In the current design, the coordination module must elicit opinions from all members at every round. A potential enhancement could entail the coordination module selecting a subset of the members for necessary conversations at later rounds, guided by the previous conversation history and proposed options.
\chedit{For example, the system can prioritize reaching out to members who have expressed uncertainty about their preferences at previous rounds. For those who have indicated their preferences firmly, the system may not need to follow up unless they later change their minds.}
This approach has direct benefits on two sides: It reduces the frequency of calls to the LLM and shortens the size of the prompt request (as it includes conversations with fewer members).
These reductions would further translate to smaller processing latency and system load for the LLMs. 
When the system is deployed on a large scale, the savings on operational costs could be substantial.

\chedit{Another challenge is that the members' preferences may change with time, and the system has to adapt to the change.} For instance, as the conversation progresses, new constraints may emerge, the priority of some constraints may shift, and certain preferences may overlap, thus presenting opportunities for consolidation. 
Therefore, the system has to track the members' preferences at different rounds so as to effectively aggregate their preferences and propose options. This functionality is not implemented in our system, and we believe that adding a preference-tracking module would enhance the system's performance. Besides, some members may attempt to manipulate the coordinator in order to bias the decision towards their favored options, e.g., paraphrasing the same preferences several times. Though there is no obvious way to detect such actions, instructing the coordinator to put an upper bound on the level to which each member's preferences should be considered can be a potential solution to this issue.

\chedit{We observed a decline in performance in our experiment when incorporating the database that includes the social graph and employees' backgrounds. While LLMs can handle structural information to some extent, they do not excel at it. Two approaches could be applied to address this challenge: augmenting LLMs with external functions that assist in managing network properties or fine-tuning LLMs for this task using more targeted data.}

Specifically, when scheduling meetings, the availability of each member’s work calendar is another practical factor to consider. 
In fact, expanding our system to incorporate this factor into the LLM’s reasoning process is not straightforward.
To examine the LLM’s ability to incorporate work calendars into its reasoning process, we conducted an experiment in which constraints based on each member’s availability were mathematically encoded into the LLM prompt for the respective member.
We observed that the LLM's performance degraded, suggesting that LLMs are less capable of dealing with hard constraints encoded in mathematical format (see, e.g., \cite{wei2022chain, rae2021scaling} and the references therein). 
The two approaches mentioned—augmenting LLMs and fine-tuning—can also be applied here to enhance the system’s ability to process calendar data. For instance, in \cite{lawless2024want}, LLMs are used to translate free-text constraints into code, which is then processed by an optimization package to identify suitable meeting times.

Finally, an issue with group recommender systems is the possibility of circular optimization, which may impact their long-term behavior (cf. \cite{dean2022preference,brantley2023ranking}). Our system aims to continuously improve the satisfaction ratio (with ties being resolved based on the satisfaction score), which indirectly improves the other metrics, as shown in the experiments. While one may think that the improvements in these metrics are expected given that a formal agreement objective is set (as in \cite{musco2018minimizing}), we argue that the improvements suggest that LLMs can follow dynamic optimization policies without hallucinations and numerical instabilities.

\subsection{Privacy Concerns}
\chedit{Our system uses private communication between each member and the system, enabling members to freely express their preferences regarding group decisions. However, there remains a concern that private information about a member—such as undisclosed preferences, sensitive attributes, or confidential details from past behaviors (potentially accessible via the database)—could be inadvertently revealed to the group when the system provides explanations for its suggestions.} Therefore, enhancing the coordination module’s capacity to protect sensitive information becomes essential to the success of the system. Such privacy concerns have gained prominence (cf. \cite{kandpal2022deduplicating,carlini2020privacy,brown2022does} and the references therein), and various strategies have been proposed to address the concern. Such mitigation strategies can include implementing LLM-based redaction and filtering modules and fine-tuning the LLMs to preserve privacy. In the context of our model, it is certainly possible to instruct the intent extraction prompts to redact any potentially sensitive information or to introduce noise in the preferred preferences (i.e., slightly perturb the reported preferences) before sending them to the coordination module.

\chedit{\subsection{LLM-simulated Users versus Human Users}
We acknowledge that human behavior cannot be fully captured by LLM-simulated users. We list several key differences as follows:
As mentioned before, humans may exhibit behavioral inconsistencies, such as changing preferences or hesitating to make definitive choices—traits that are difficult to simulate accurately with LLMs.
Real users bring emotional and psychological factors into group decision-making, including motivation, fatigue, social influence, and stress under time constraints.
Additionally, human users may have privacy and trust concerns, particularly if they worry that sensitive information could be exposed or misinterpreted by the system. This may affect their willingness to share authentic preferences, a factor typically absent in LLM simulations.
Finally, human users may employ strategies to influence group decisions to their advantage, an aspect commonly observed in social interactions. For example, they may misrepresent their preferences to prompt the coordinator to suggest a more favorable option. LLMs struggle to capture such complex, strategic thinking.}

\subsection{A New Way of Work Coordination}

Our system exhibits the potential to facilitate collective decision-making by coordinating conversations between members. It actively understands members’ preferences and suggests alternatives that aid in reaching a consensus. \chedit{Our case study of meeting scheduling can be extended to other working contexts, such as work plan formulation and idea brainstorming. This requires adaptation of LLM promoting for a working scenario of interest.} The use of LLMs can enhance productivity and creativity by expediting the decision-making process through real-time analysis of vast data and presentation of relevant insights. Furthermore, LLMs consider multiple perspectives, ensuring all relevant viewpoints are taken into account for more inclusive and comprehensive decision-making. Beyond facilitating collective decision-making, LLMs can significantly boost productivity by alleviating non-essential work, thereby saving time and effort and reducing costs.

\section{Literature Review} \label{sec:related_work}

\subsection{Collective Decision-Making in Work Environments}

There has been a significant amount of work regarding collective decision-making tasks in work environments. In particular, meeting scheduling has been a focus scenario for collective decision-making due to its high complexity and ubiquity in various social and organizational settings \cite{crawford2007experts,crawford2009learning,ephrati1994meet,benhassine2007agent,brzozowski2006grouptime,mok2023challenging}.

The growing trend of organizations becoming more distributed, coupled with the necessity for collaboration among employees in various locations, has led to a wide range of free-text preferences and organizational asymmetries \cite{mok2023challenging,reinecke2013doodle}.
The trend prompts the development of distributed meeting scheduling assistants, such as \emph{Calendar.help} \cite{cranshaw2017calendar}.
However, as mentioned in \cref{sec:requirements_analysis}, \emph{Calendar.help} considers workers' calendars and cannot handle free-text preferences, whereas our system can identify meeting schedules that balance diverse free-text preferences among several members, considering social relations and organizational asymmetries.

Furthermore, having little time for focus work has been a major problem for information workers nowadays. The work of \cite{koutsuv2023focus} shows that tools that assist in scheduling focus time have positive effects on the well-being of workers. Our findings can be seen as complementary to this work, as our system can help alleviate non-essential work (e.g., meeting scheduling) and accelerate collective decision-making tasks (e.g., brainstorming) by leveraging LLMs.

\subsection{Understanding Human Preferences Using Large Language Models} 

Large language models have been utilized in many studies to comprehend human preferences. These models analyze historical interactions and human feedback to discern preferences for specific topics, products, or services. A notable application of LLMs is their use in tasks such as collaborative filtering to understand human preferences.  The research by \cite{kang2023llms} explores whether LLMs can learn to rank similarly to ranking algorithms in a zero-shot or few-shot manner and shows that LLMs can achieve performance akin to collaborative filtering algorithms with substantially less data. Similarly, in \cite{sanner2023large}, it is shown that zero-shot and few-shot-trained LLMs are able to perform competitively compared to state-of-the-art cold-start collaborative filtering methods for recommendations from language and item-based preferences, offering additional explainability over traditional methods. Our work contributes to this field of work as our system extracts group members' individual preferences from their dialogues with the system and accordingly makes suggestions for the group. 

Another development is using machine learning methods to generate consensus statements by reconciling various opinions. The work of \cite{cao2018attentive} introduces AGREE, which is a neural collaborative filtering method for groups and users by utilizing a learned preference aggregation mechanism based on attention. Compared to AGREE, our method is also able to aggregate different preferences; however, in our case, we do not specifically train models for handling preferences, and instead, we utilize LLMs to consolidate them. 
Besides, in \cite{bakker2022fine}, the authors fine-tune LLMs to aggregate members' opinions and generate consensus statements. They train reward models to predict individual preferences, enabling them to quantify and rank consensus statements according to different social choice functions. Our work contributes to this development as it also leverages LLMs to find consensus, balancing the opinions of multiple members.

\subsection{Simulating Human Behavior Using Large Language Models}

Using LLMs to simulate human behavior in user studies has emerged as an innovative approach, offering both efficiency and versatility. LLMs, using extensive training data and contextual understanding, can effectively emulate the responses and behaviors of a diverse range of humans, covering a broad spectrum of demographics and perspectives. This simulation allows researchers to accelerate data collection, streamline experimentation processes, and explore numerous hypothetical scenarios without the logistical challenges of recruiting and managing actual human participants. Moreover, the adaptive nature of LLMs enables researchers to readily adjust the characteristics of the simulated personas, ensuring a variety of plausible reactions in line with our paper's objectives. 

Particularly, simulating agents with human-like behavior using LLMs has garnered significant interest in the field of social sciences \cite{grossmann2023ai}. For instance, LLM-simulated agents have been employed to study social behavior \cite{park2023generative}, replicate human subject studies \cite{aher2022using}, perform market research \cite{brand2023using}, act as economic agents \cite{horton2023large}, conduct qualitative interviews at a large scale \cite{chopra2023conducting}, and perform crowdsourcing tasks \cite{veselovsky2023artificial}. Our paper contributes to this line of study by utilizing LLM-simulated members to test and evaluate our system, addressing the challenges and costs associated with conducting user studies involving human participants (e.g., the length of 18 months for the user study of \emph{Calendar.help} \cite{cranshaw2017calendar}).

\subsection{Evaluating the Performance of Large Language Models}
As the application of LLMs becomes increasingly widespread across various domains, gaining a comprehensive understanding of their capabilities and limitations has become critical for researchers and developers.
Researchers have introduced diverse evaluation frameworks, including the capabilities of LLMs in natural language understanding and generation, human-AI interaction assessments, benchmarking on novel scenarios, and assessing ethical considerations such as fairness and bias. 

Particularly, in \cite{liang2022holistic}, the authors introduce  Holistic Evaluation of Language Models (HELM). HELM follows a three-step process. First, it categorizes various potential use cases and evaluation metrics for LLMs, highlighting gaps like underrepresented dialects and trustworthiness measures. Second, HELM adopts a multi-metric approach, assessing seven metrics across multiple scenarios to ensure a comprehensive understanding of LLM performance. It also carries out seven specific evaluations targeting areas such as reasoning and disinformation. Third, HELM conducts a large-scale evaluation of multiple LLMs on multiple scenarios with standardized conditions.

On the other hand, the paper \cite{lee2022evaluating} proposes an evaluation framework known as  Human-AI Language-based Interaction Evaluation (HALIE).
HALIE emphasizes the interactive process rather than just the final outcome, takes into account the user’s personal experience, and extends its assessment beyond quality to include factors such as enjoyment and ownership.
The authors introduce five tasks that span various types of interaction and evaluate them using four different LLMs. The authors conclude that superior performance in non-interactive tasks does not necessarily translate into better human-LLM interaction. In a similar vein to our work, HALIE underscores the importance of incorporating human feedback through qualitative evaluation, in addition to a quantitative assessment of the system, to effectively measure the system’s reasoning capabilities.

Both HELM and HALIE are primarily designed to evaluate the capabilities of LLMs exclusively. In contrast, as our research is focused on creating a work coordination system,  our primary emphasis is on establishing metrics for assessing coordination capabilities and devising a comprehensive method to evaluate the coordination process at a large scale. Our evaluation approach includes conducting a large-scale experiment that uses LLMs to simulate discussions, complemented by a smaller-scale study involving human participants.



\subsection{AI and Collective Decision-Making} 
Recent research has explored how artificial intelligence can assist in collaborative decision-making, with roots tracing back to graph mining techniques, see, e.g., \citep{zhang2018characterizing,gu2021case}. The work of \cite{bakker2022fine} demonstrates that fine-tuning LLMs can generate statements that maximize the expected approval from a diverse group of people with diverse opinions. Our research contributes to the growing area of using machine learning methods for decision-making (cf. \cite{landemore2020open,fishkin2009people}). 

Moreover, there has been a focus on developing communicative agents. These agents are constructed to effectively engage and communicate with other agents or human users using natural language processing techniques such as LLMs. The main objective is to enhance interaction and collaboration efficiency, allowing various agents to communicate and negotiate to address more complex tasks autonomously. Various studies have explored different aspects of communicative agents, proposing frameworks like cooperative agent models for role-playing and creating controlled environments with virtual entities capable of emulating human behavior \cite{park2023generative}. 
Researchers also introduce chat-based software development frameworks \cite{qian2023communicative,nguyen2018chat}, systems for negotiating preferences \cite{alvarez2016hootle+}, curated datasets aligned with human preferences \cite{liu2023training}, and a multi-agent debate framework to improve results in diverse scenarios such as translation, arithmetic, and text evaluation problems \cite{liang2023encouraging,du2023improving,chan2023chateval,li2023prd,wu2023autogen}. Our work contributes to this body of work as we study collective decision-making and assess our framework by simulating agents with LLMs.

\section{Conclusion} \label{sec:conclusion}

In this paper, we developed a system that leveraged LLMs for collective decision-making. We provided the general framework for group decision-making and showed how our framework can particularly be used in group meeting scheduling, a widespread task that involves coordinating multiple members with different preferences and opinions to find a mutually agreeable time slot. We found out that our system was able to propose options that increasingly fulfilled a larger number of members' preferences in a balanced way through a large-scale study involving members simulated by LLMs. Moreover, our system was found to possess robust skills in preference aggregation and reasoning, as reported by recruited human participants. Our proposed system can have many potential applications that require the solicitation and reconciliation of preferences and opinions, such as mediating team discussions, brainstorming, and work plan creation, opening new pathways for the future of work and productivity. Finally, our work showcases the use of LLMs to simulate human participants in large-scale studies that would otherwise be very costly and time-consuming to perform.

\section{Ethics Statement} \label{sec:ethics_statement}

Due to our evaluation involving participants assessing the system's ability to exhibit human-like reasoning (RQ4), we are cognizant of potential privacy concerns for the individuals in our study. To address these concerns, our research underwent a formal privacy review within our organization, ensuring that individual employees' identities remain protected. Additionally, all survey participants provided explicit consent for us to use their responses in our research, including this paper. You can find the consent form used for our study in \cref{app:user_study}. We have taken measures to anonymize all results by removing any names or email addresses and presenting responses at a higher-level aggregation. Individual data has not been analyzed or disclosed in this study.

Furthermore, we have omitted potentially sensitive metadata, such as indicators of seniority and job function. To maintain confidentiality, access to the dataset has been limited to the research team involved in this project. Participants were compensated for their participation in the user study.

\begin{acks}
We want to thank the researchers at Microsoft's Office of Applied Research -- particularly Bahar Sarrafzadeh, Victor Poznanski, Yujin Kim, Katrina Zuccaro, and Aaron Halfaker -- for stimulating discussions and feedback on our project. We would also like to thank Shilad Sen for providing us with the synthetic dataset for meeting scheduling. Finally, M.P. would like to thank Jon Kleinberg and Farhana Shahid for discussing the early versions of the manuscript.  
\end{acks}

\bibliographystyle{ACM-Reference-Format}
\bibliography{references}

\newpage

\appendix

\begin{center}
\Huge
    \textbf{Appendix}
\end{center}

\begin{center}
\Large
    \textbf{Part I: Synthetic Data and Prompts for Meeting Scheduling}
\end{center}

\section{Example Data} \label{app:data_specifications}

\subsection{Member Data}

\begin{table}[H]
\centering
\footnotesize
\caption{Example Synthetic Member Data.} \label{tab:user_data}
\begin{tabular}{p{0.1\textwidth}p{0.1\textwidth}p{0.1\textwidth}rp{0.25\textwidth}p{0.25\textwidth}}
\toprule
Name & Manager & Role & Level & Responsibilities & Schedule Preferences \\
\midrule
Member 1 &  & CEO & 5 & ['Setting the vision, mission, and values of the company', "Leading the overall strategy and execution of the company's initiatives", 'Managing the financial, legal, and operational aspects of the company'] & She prefers to have meetings in the morning and block out time for deep work in the afternoon. She is flexible with her availability but appreciates advance notice and clear agendas. She uses a shared calendar to keep track of her appointments and tasks. \\
Member 34 & Member 1 & Senior Customer Success Manager & 3 & ["Managing a portfolio of key accounts and ensuring they achieve their desired outcomes and satisfaction with the company's products and services", 'Providing proactive and reactive support, guidance, and training to customers', 'Identifying and pursuing opportunities to upsell, cross-sell, and renew customers'] & He likes to have meetings throughout the day and adjust his schedule according to his customers' preferences and availability. He is flexible with his availability, but prefers to have meetings scheduled at least a day in advance. He uses a customer success platform to manage his customer interactions and outcomes. \\
Member 33 & Member 1 & Senior Software Engineer & 3 & ["Designing, developing, and testing software solutions for the company's products and services", 'Troubleshooting and resolving software issues and bugs', 'Mentoring and coaching junior software engineers and sharing best practices'] & He prefers to have meetings in the morning and use the rest of the day for coding and debugging. He is flexible with his availability, but prefers to have meetings scheduled at least a few hours in advance. He uses a code review tool to collaborate and communicate with his peers and managers. \\
Member 29 & Member 22 & Sales Representative & 2 & ["Identifying and qualifying leads and prospects for the company's products and services", "Conducting sales presentations and demos to potential customers and explaining the value proposition and benefits of the company's products and services", 'Following up and closing sales deals and ensuring customer satisfaction and retention'] & She likes to have meetings in the afternoon and evening, and use the morning and early afternoon for prospecting and following up with leads and customers. She is flexible with her availability, but likes to have meetings scheduled at least a day in advance. She uses a CRM tool to manage her sales pipeline and activities. \\
\bottomrule
\end{tabular}
\end{table}

\subsection{Meeting Data}

\begin{table}[H]
    \centering
    \footnotesize
    \caption{Example Synthetic Meeting Data.}
    \begin{tabular}{lp{0.3\textwidth}lll}
    \toprule
    Organizer & Members & Subject & Date & Duration \\
    \midrule
    Member 34 & Member 1, Member 29, Member 33 & Discuss Customer Success Efforts & 2/16/2023 & 30 minutes \\
     Member 14 & Member 13, Member 33, Member 15, Member 4, Member 11 & AI feature update and feedback & 2/15/2023 & 30 minutes \\
    \bottomrule
    \end{tabular}
    \label{tab:meeting_data}

\end{table}

\subsection{Social Graph Data}
\begin{figure}[t]
    \centering
    \subfigure[The social relations of our synthetic company  \label{subfig:company_network}]{\includegraphics[width=0.45\textwidth]{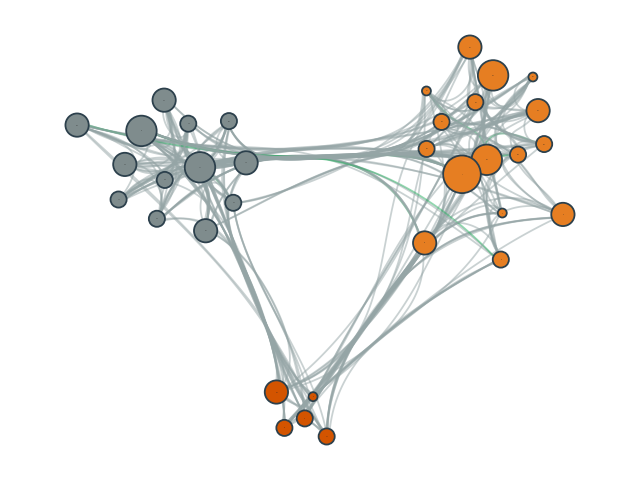}}
    \subfigure[The subgraph  corresponding to a meeting scenario \label{subfig:meeting_network}]{\includegraphics[width=0.45\textwidth]{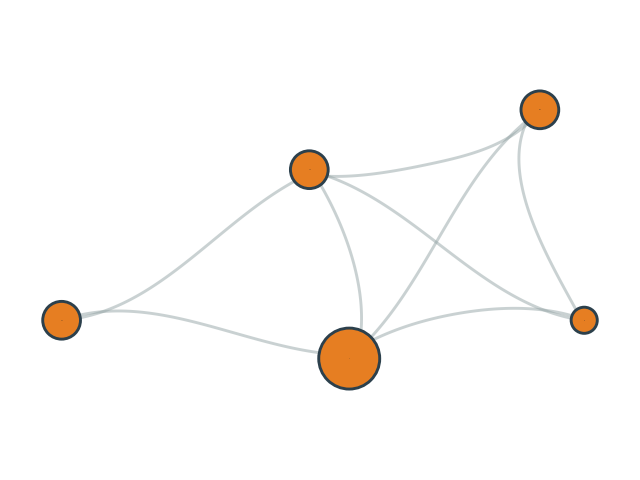}}
    
    \caption{The size of each node denotes seniority level. The collaborator relations are drawn in gray color, and the managerial relations are drawn in green color. The nodes are clustered in communities. Example synthetic user metadata is given in \cref{tab:meeting_data}.}
    \label{fig:company_network}
\end{figure}

\cref{fig:company_network} shows an illustration of the company network. 
\cref{tab:social_graph_meeting} contains an example induced social graph. 

\begin{table}[H]
\footnotesize
\caption{Social Graph of \cref{subfig:meeting_network}.} \label{tab:social_graph_meeting}
\begin{tabular}{lll}
\toprule
 Source member & Target member & Relation \\
\midrule
Member 4 & Member 34 & collaborator \\
Member 4 & Member 1 & collaborator \\
Member 4 & Member 13 & collaborator \\
Member 34 & Member 1 & collaborator \\
Member 34 & Member 13 & collaborator \\
Member 12 & Member 1 & collaborator \\
Member 12 & Member 13 & collaborator \\
Member 1 & Member 13 & collaborator \\
\bottomrule
\end{tabular}    
\end{table}

\section{Prompt Design for Group Meeting Scheduling} \label{app:prompts}

We give all of the prompts used for the experiments. We follow the Python \texttt{f-string}\footnote{\url{https://docs.python.org/3/tutorial/inputoutput.html\#formatted-string-literals}} format. The prompts are suitable for the \texttt{gpt-3.5-turbo} model. 

\subsection{Intent Extraction Prompts} \label{app:solicitor_prompts}

We present the intent extraction module for round $2 \le t \le T$. The intent extraction module consists of two prompts: 

The first prompt instructs the LLM how to communicate with member $i$ at any round $t$. The prompt is given below:

\begin{lstlisting}
<|im_start|>system
# Goal
You are an AI assistant for a user and your goal is to aid the user to express their preferences.

# Task
You are given information about a message and some suggested options and your goal is to aid the user to agree with one of these options.
Specifically, you should obey to the following instructions:
1. You should start the conversation by listing all the suggested options and the reasons behind them.
2. If the user does not agree with any option, you should ask the user to provide at least one reason explaining why these options do not work for them and ask them to provide alternative options.
3. When the user agrees with an option or suggests an alternative, you must output <EXIT>.
4. You only support aiding the user to decide if the existing preferences work or not and suggesting alternatives, not small talk.
5. Do not fabricate any data. Your answer MUST be based on the responses of the users you are talking to.

# Message
The message is given below after chevrons: 

<MESSAGE>
{message}
</MESSAGE>

# OPTIONS
The suggested options are given below after chevrons in JSON format:

<OPTIONS>
{coordinator_options}
</OPTIONS>

# Notes
* You should be concise to your answers to the user.
* You should never mention that you are an AI assistant.
* Your answers should be based only on the data provided above.
* When the user agrees with an option or suggests an alternative, you must output <EXIT>.

<|im_end|>
\end{lstlisting}

Here \texttt{organizer\_message} corresponds to the organizer message, and \texttt{coordination\_results} corresponds to the JSON-serialized options $\cX^{t - 1}$ suggested by the organizer at round $t - 1$.

The second prompt corresponds to summarizing the conversation $m_i^t$ and extracting the intents $\theta_i^t$ for each member $i$ at round $t$:

\begin{lstlisting}
<|im_start|>system
# Task
You are given the chat history between {member.name} and an AI assistant.
Your job is to summarize the conversation between {member.name} and the AI assistant.

You MUST output your answer in JSON format with the following structure:

{
    "preferences" : list of the preferences of {member.name},
    "option" : if the user has agreed with an option list it here,
}


# Chat history
The chat history between the user and the AI assistant is given below after chevrons:

<CHAT HISTORY>
{chat_history}
</CHAT HISTORY>

# Notes
* You should not fabricate any options or preferences. All of your answers should be based on the chat history provided.
* You MUST output JSON only and nothing else. 

<|im_end|>
\end{lstlisting}

In this prompt, \texttt{chat\_history} is the chat history $m_i^t$ between a member and the intent extraction module, and \texttt{member.name} is the name of the corresponding member. 

For $t = 1$, the prompt is similar, with the only difference being that it omits the part referring to $\cX^{t - 1}$. 

\subsection{Coordination Module Prompt} \label{app:coordinator_prompts}

We present the coordination module prompt for rounds $2 \le t \le T$, which corresponds to the most general setting:

\begin{lstlisting}

<|im_start|>system
# Task
You are an AI assistant and your goal is to aid users to reconcile their options and suggest meeting times that work for most of the users based on their preferences. 
You are given information about the preferences of the users and your goal is to determine if there there is a common ground between the users. If the preferences of the users diverge, you should suggest at most {K} meeting times that reconciliate the preferences of at least 2 users.

1. You must always list the candidate option provided below.  
2. You should suggest at most {K - 1} options that is different than the candidate option and can satisfy at least {num_candidate_users} users.
3. If the organizer mentions that some user is needed to attend, then at least one option must work for this user.
4. You should prefer options where the attendees are collaborators or teammates by using information from the relationships between the employees.
5. You should prefer options that work both for users and their managers by using information from the relationships of the employees. 

# Candidate Option
The candidate option is given below after chevrons:

<CANDIDATE OPTION>
{candidate_option}
</CANDIDATE OPTION>

# Preferences
The preferences of the users regarding the meeting time are given below after chevrons:

<PREFERENCES>
{summarized_preferences}
</PREFERENCES> 

# Message
In your results, you should consider any important information described in the message and any constraints and preferences it outlines. 
The message is given below after chevrons:

<MESSAGE>
{organizer_message}
</MESSAGE>

# Relationships
You should take into account the relationships of the employees. 
The relationships between the attendees of the meeting are given below after chevrons in JSON format:

<RELATIONSHIPS>
{relationships}
</RELATIONSHIPS>

# Roles And Responsibilities

If the meeting invite requires attendees with specific roles and responsibilities to be in the meeting, then you should take into account the roles and responsibilities of these attendees.
The roles and responsibilities of the attendees of the meeting are given below after chevrons:

<ROLES>
{roles}
</ROLES>

<RESPONSIBILITIES>
{responsibilities}
</RESPONSIBILITIES>



# Output Instructions
You MUST produce your output in JSON format. Do not output anything else.
The structure of the JSON is given below:

{
    option1 : {
        "option" : the first option,
        "users": the users for whom this option1 works,
        "reasons" : list of reasons why option1 works,
    },
    option2 : {
        "option" : the second option,
        "users": the users for whom this option2 works,
        "reasons" : list of reasons why option2 works,
    }, ...
}


# Notes
* You should be strategic, your goal is to help users reach consensus.
* You should only provide reasons that the users have expressed. Do not fabricate your own reasons.
* You MUST output JSON only and nothing else.
<|im_end|>

\end{lstlisting}

Here, the variable \texttt{summarized\_preferences} corresponds to a dictionary with key-value pairs $(i, \theta_i^t)$ for every member $i \in [n]$.  \texttt{candidate\_option} corresponds to the decision candidate $\hat x^{t - 1}$, \texttt{num\_candidate\_users} correspond to the number of members for whom $\hat x^{t - 1}$ satisfies at least one preference (i.e. $n \hat r^{t - 1}$), \texttt{organizer\_message} corresponds to the meeting invite, \texttt{relationships} corresponds to the JSON-serialized social graph between the meeting members, \texttt{roles} (resp. \texttt{responsibilities}) corresponds to the JSON-serialized roles (resp. JSON-serialized responsibilities) of the members, and \texttt{K} corresponds to the maximum number of suggested options. This prompt is used to produce $\cX^t$ in JSON format.

\subsection{Evaluation Prompt} \label{app:scoring_prompt}

We present the evaluation prompt (for any round):

\begin{lstlisting}

<|im_start|>system
# Task
You are given information about users and their expressed preferences as well as a set of available options
and your job is to provide a score ranging from 1 to 3 for each user and each option. 

## Output Format
Your output must be formatted in JSON with the following structure:
{
"option": option name,
"scores" : [
{
    "user" : user name, 
    "score" : integer score ranging from 1 to 3 that the user provides for this option (instructions are provided below),
    "reasons" : list of reasons for providing the chosen score
},...],
...
}

## Score Values
The score values have the following meaning:
1: the option satisfies few of the preferences of the user
2: the option satisfies at least half but not all the preferences of the user
3: the option satisfies all of the preferences of the user 

## User Preferences 
The user preferences are provided below after chevrons formatted as JSON:

<USER PREFERENCES>
{summarized_preferences}
</USER PREFERENCES>

## Available Options
The available options are provided below after chevrons formatted as JSON:

<AVAILABLE OPTIONS>
{coordinator_options}
</AVAILABLE OPTIONS>

# Notes
* Do not fabricate any answers. Your answers should only be based on the available data.
* You MUST output JSON only and nothing else.

<|im_end|>

\end{lstlisting}

Here, the variable \texttt{summarized\_preferences} corresponds to a dictionary with key-value pairs $(i, \theta_i^t)$ for every member $i \in [n]$, and the variable \texttt{coordinator\_options} corresponds to the JSON-serialized option set $\cX^t$. This prompt produces the scores $\{ \pi_{j, i}^t \}_{j \in [K], i \in [n]}$.

\subsection{Synthetic Member Prompts} \label{app:user_prompts}

We present the member prompt that is used to simulate an artificial member. Here, the fields have the following meanings:

\begin{itemize}
    \item \texttt{member.name}: The member's name.
    \item \texttt{member.role}: The member's role (e.g., Software Engineer).
    \item \texttt{member.manager}: The member's manager.
    \item \texttt{member.teammates}: List of the member's teammates.
    \item \texttt{member.collaborators}: List of the member's collaborators.
    \item \texttt{member.preferences}: List of the member's schedule preferences.
\end{itemize}

\begin{lstlisting}
<|im_start|>system

# User Profile
Your name is {member.name} and you work as {member.role} at {company}. You report to {member.manager}. Your teammates are {', '.join(member.teammates)}. Your collaborators are {', '.join(member.collaborators)}. 

You use the AI assistant to find a time for hosting a meeting.
Your job is to
1. Express your preferences to the AI assistant 
2. Respond to any clarifying question made by the assistant

# Preferences
Your  preferences written in third-person are the following:
{'\n'.join(member.preferences)}

# Notes:
- Always use the "I" pronoun when replying.  
- Never mention your responsibilities as part of your response.
- Your goal is to find a feasible time for your meeting. You MUST not engage in small-talk.

<|im_end|>
\end{lstlisting}

\section{Simulation Plots} \label{app:coordinator_results}

We present detailed plots from simulations. The aggregated results of these plots are presented in \cref{fig:trend_lines_results_heterogeneous,fig:trend_lines_results_homogeneous,fig:trend_lines_results_homogeneous_options}. 

\begin{figure}[H]
    \centering
    \includegraphics[width=0.7\textwidth]{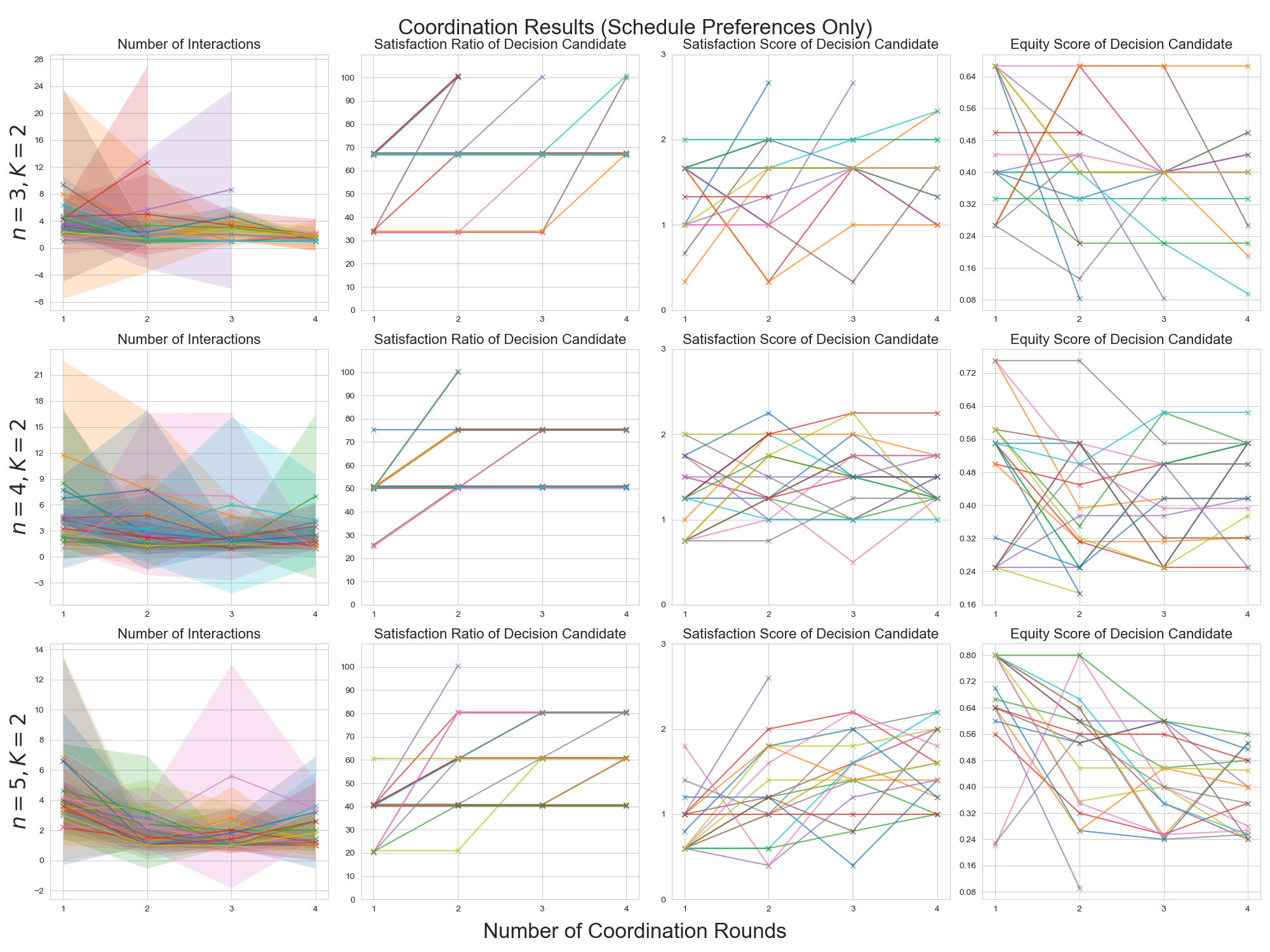}
    \caption{Results from simulations on the baseline setting.}
    \label{fig:majority_results_homogeneous}
\end{figure}

\vspace{-1em}

\begin{figure}[H]
    \centering
    \includegraphics[width=0.7\textwidth]{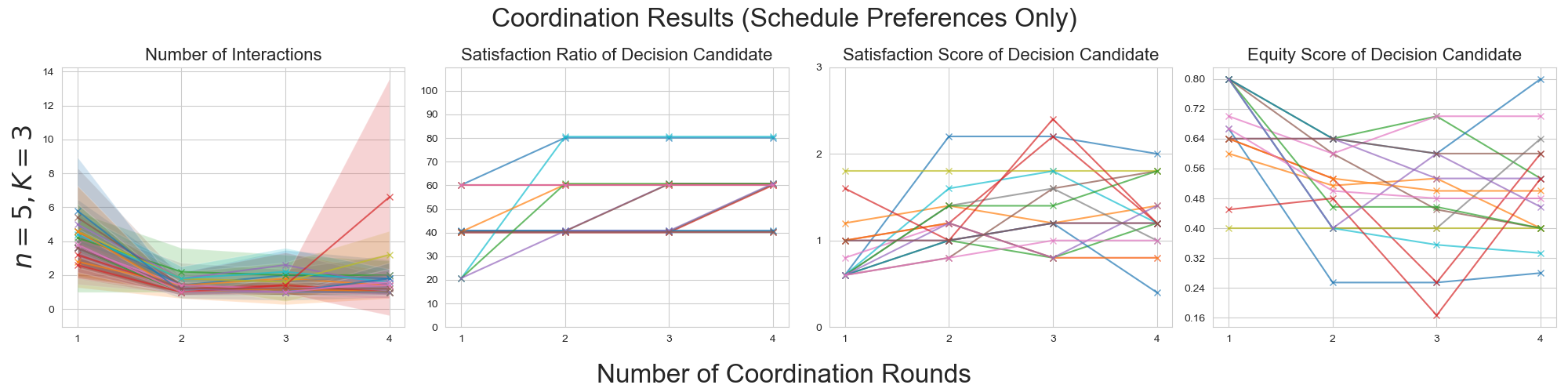}
    \includegraphics[width=0.7\textwidth]{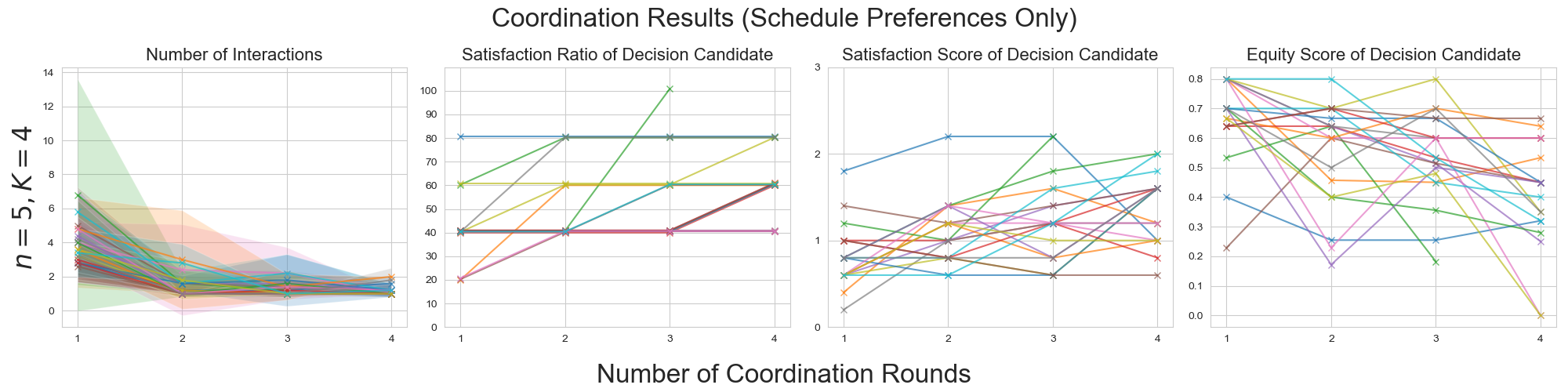}
    \caption{Results from simulations on the baseline setting where the number of options $K$ varies.}
    \label{fig:majority_results_homogeneous_options}
\end{figure}

\vspace{-1em}

\begin{figure}[H]
    \centering
    \includegraphics[width=0.7\textwidth]{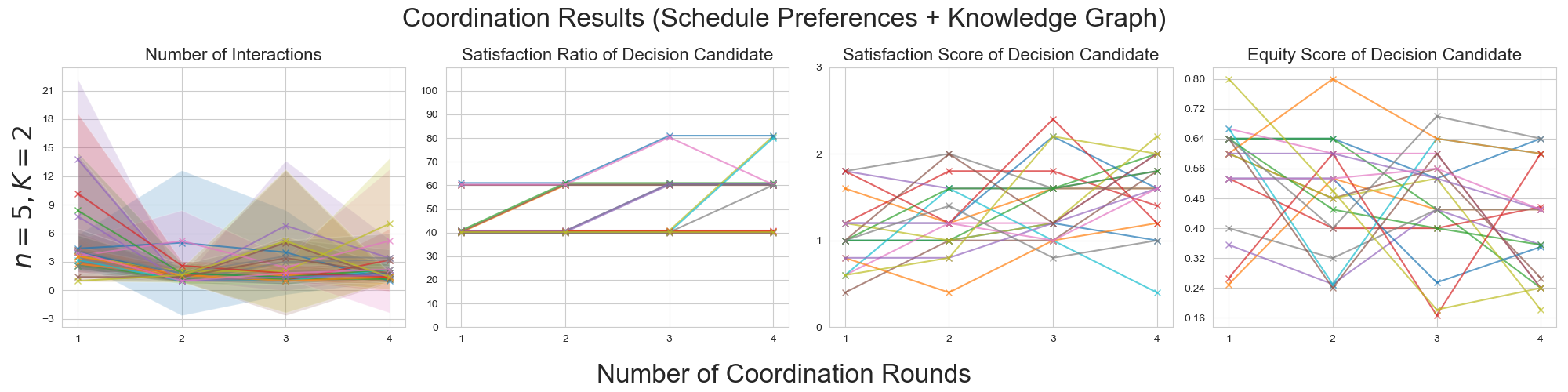}
    \caption{Results from simulations from the study where the schedule preferences and the database are taken into account.}
    \label{fig:majority_results_heterogeneous}
\end{figure}

\newpage

\begin{center}
\Large
    \textbf{Part II: System Evaluation with Human Participants}
\end{center}

\section{Survey} \label{app:user_study}

\begin{figure}[t]
    \centering
    \includegraphics[width=0.4\textwidth]{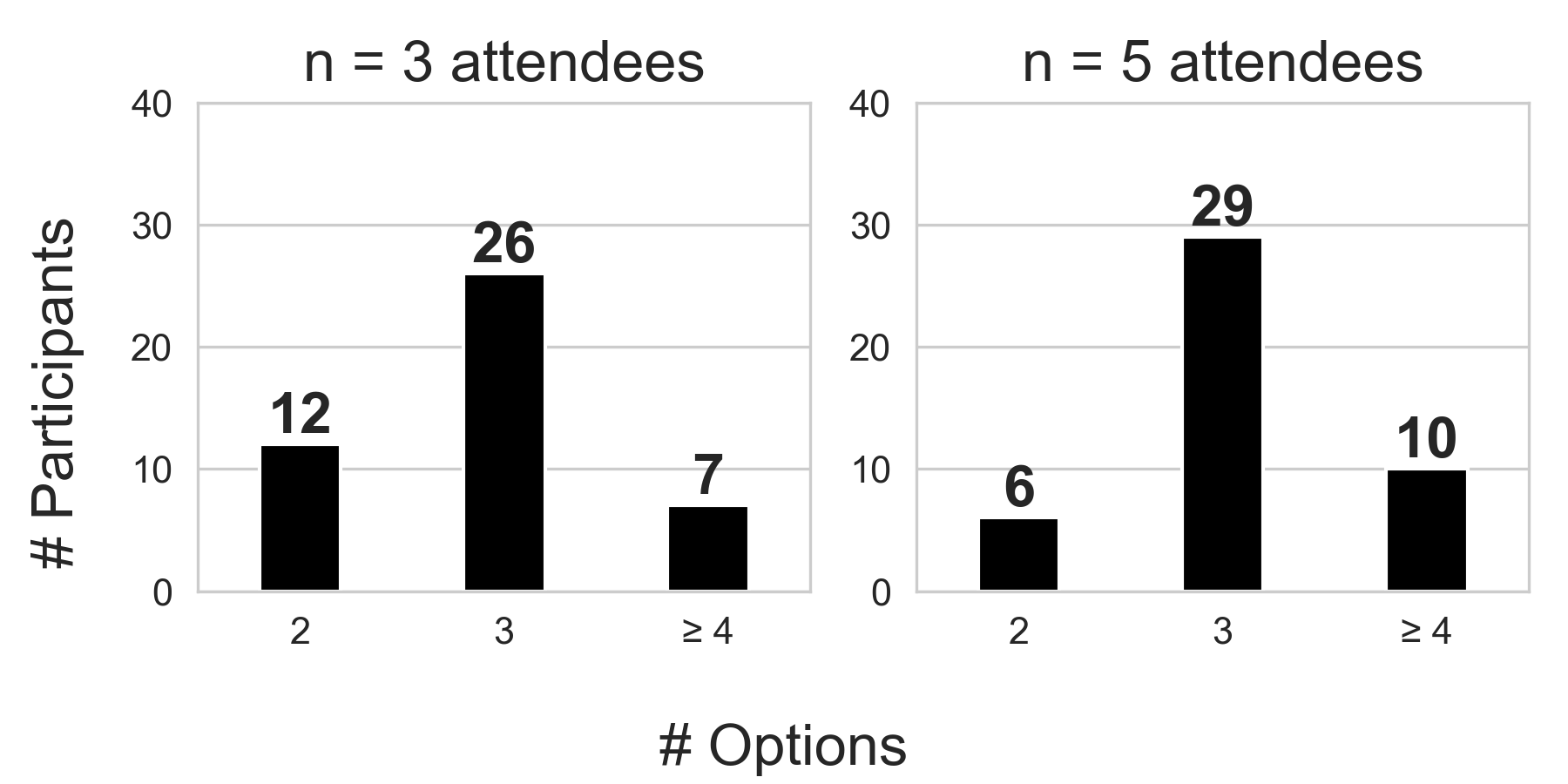}
    \caption{The participants' preferred number of options for meetings with $n = 3$ and $n = 5$ members.}
    \label{fig:number_of_options_user_study}
\end{figure}

\subsection{Consent and Intake Forms} 

\noindent \textbf{Procedures:} As part of this study, you are expected to provide a qualitative evaluation of our system. Specifically, you will examine the experimental results for up to 3 scenarios, each involving diverse individual preferences on scheduling and a coordination AI-agent that proposes results. Your evaluation will help us to assess our system. In this study, you will complete:

\begin{enumerate}
    \item A 2-minute intake survey that solicits your demographic info, how large and how many meetings you have per day, and any collaborative tools that you may use.
    \item Evaluation of up to 2 scenarios. For each evaluation, you will need to provide responses involving qualitative feedback (e.g., "To what degree do you believe the system takes into account the users' preferences?", "Are the reasons provided by the system acceptable?", etc.).
    \item Finally, you will be asked to reply to some questions regarding the need for a coordination agent. In total, your participation will take between 25 to 35 minutes.

\end{enumerate}

The personally identifiable data, such as your name and email, will only be used for study coordination purposes. We will delete this personally identifiable data within 12 months. The study information we collect during the study will be de-identified and only associated with a participant code, which we provide to you when you enroll in the study. We plan to store this de-identified and anonymized study data indefinitely for future studies relating to scheduling.

\smallskip

\noindent \textbf{Benefits:} You will be compensated for joining this study. We hope that the findings from this research will support the design of an intelligent assistant for collective decision-making.

\smallskip

\noindent \textbf{Risks:} The risks of participating are similar to what you might experience while performing everyday tasks. Risks include fatigue and frustrations due to additional time and effort required to fill out the survey entries or accidental disclosure of personally identifiable information. We remind you not to disclose personally identifiable information in all of the open text fields. In the case of accidental disclosure, you may contact the study team to remove the sensitive information, and the study team will scrub the data to ensure it is properly anonymized.

\smallskip

\noindent \textbf{Privacy \& Confidentiality:} Researchers will keep your participation and the information you share as confidential as possible. The information you share will be labeled in our records with a code instead of your name or other direct identifier. The key to this code will be stored separately and destroyed after 12 months. The de-identified and anonymized study data will be retained indefinitely to support future studies related to work coordination. Researchers may share the results of this study publicly, such as in journal articles or conference presentations, but your identity will not be disclosed. Information collected during this study may be used for future research studies or to improve products or services. If that happens, researchers will remove any direct identifiers, like your name or email address, before sharing.

If you decide to withdraw from the study and want researchers to remove your study information, you can contact us. However, after we remove any link to identifiers, it would no longer be possible to delete your data, but any results of the research will not identify you individually.

\smallskip

\noindent \textbf{Participation is your choice:} Whether or not you participate is entirely up to you. You can decide to participate now and stop participating later. Your decision of whether or not to participate will have no impact on any other services or agreements you have with our organization outside of this research.

\smallskip

\noindent \textbf{Questions or Concerns:} If you have any questions or concerns about this study at any time, you may contact us. Should you have any questions about your rights as a research subject, please contact us.

By clicking ``I agree'' below, you confirm that the study was explained to you, you had a chance to ask questions before beginning the study, and all your questions were answered satisfactorily. By clicking ``I agree'' below, you voluntarily consent to participate, and you do not give up any legal rights you have as a study participant. You may request a link to download this form. On behalf of our organization, we thank you for contributing to our research.

\begin{itemize}
    \item I agree. I consent to participate in this study. 
    \item No thanks, I would not like to participate in this study
\end{itemize}

\smallskip

\subsection{Eligibility Criteria}

The participants must be at least 18 years old, based in the United States,  not belong to a sensitive group (as defined in our organization), and have experience in using meeting scheduling tools.

\subsection{Requirements Analysis Form} Below, we give the questions we asked for the requirements analysis form. The responses are summarized in \cref{fig:demographics}.

\begin{itemize}
    \item What is your role in the company? (Free-text)\footnote{We do not share the results of this question in the paper to mitigate privacy risks.}
    \item How many years of experience do you have in that role or similar roles? (Multiple Choice)

    \emph{Choices: Less than 1 year, 1 -- 3 years, 1 -- 6 years, 6 -- 10 years, 10 years or longer}

    \item On average, how many members do your meetings contain? 
        
    \emph{Choices: 3 to 5, 6 to 10, 10 or more}

    \item On average, how many meetings do you have per day? 
    
    \emph{Choices: 1 to 2, 3 to 5, 6 or more}

    \item In which of the following scenarios would you be interested in having an LLM-based copilot that assists with collective decision-making?

    \emph{Multiple Choice (can select more than one choice): Meeting Scheduling, Group Discussion, Brainstorming, Event Planning, Other (free-text)}

\end{itemize}

\subsection{Synthetic Meeting Scenario}

John Doe organized the meeting, and the meeting members were Norma Fisher, Elizabeth Woods, and Theodore Mcgrath.

\smallskip

\noindent \textbf{Meeting Invite.} The meeting invite follows:

\begin{quote}
\itshape
Hi everyone, 

\smallskip

I hope this message finds you well. I wanted to reach out to you today to discuss some important matters regarding our customer success efforts. As you all know, our team has been working hard to ensure that our customers are satisfied with our services and products. I would like to invite you to a meeting on February 16, 2023, which will last for 30 minutes. I would appreciate it if you could let me know what time works best for you. Additionally, I would like to know if there are any specific topics or concerns that you would like to discuss during the meeting.

\smallskip

Looking forward to hearing back from you soon. 

\smallskip

Best regards,

John Doe
\end{quote}

\smallskip

\noindent \textbf{Schedule Preferences.} The schedule preferences of the members follow:

\begin{itemize}
    \item \emph{Norma Fisher.} She prefers to have meetings in the morning and block out time for deep work in the afternoon. She is flexible with her availability but appreciates advance notice and clear agendas. She uses a shared calendar to keep track of her appointments and tasks.

    \item \emph{Elizabeth Woods.} She prefers to have meetings in the middle of the day and leave the mornings and afternoons for focused work. She is flexible with her availability but prefers to have meetings scheduled at least a day in advance. She uses a project management tool to keep track of her product development tasks and progress.

    \item \emph{Theodore Mcgrath.} He likes to have meetings throughout the day and adjust his schedule according to his customers' preferences and availability. He is flexible with his availability but prefers to have meetings scheduled at least a day in advance. He uses a customer success platform to manage his customer interactions and outcomes.
\end{itemize}

\smallskip 

\noindent \textbf{Proposed Options by our System.} The proposed options by our system for round $t = 1$ follow: 

\begin{enumerate}
    \item February 16, 2023 at 10 am EST
    \begin{itemize}
        \item \emph{Satisfies:} Norma Fisher, Theodore Mcgrath
        \item \emph{Reasons:} 
        \begin{enumerate}
            \item Norma Fisher is available anytime between 9 am and 12 pm EST for the meeting on February 16, 2023.
            \item Theodore Mcgrath is available all day on February 16, 2023.
        \end{enumerate}
    \end{itemize}
    \item February 16, 2023 at 2pm EST
    \begin{itemize}
        \item \emph{Satisfies:} Norma Fisher, Elizabeth Woods
        \item \emph{Reasons:} 
        \begin{enumerate}
            \item Norma Fisher prefers to have meetings in the morning and block out time for deep work in the afternoon.
            \item Elizabeth Woods prefers meetings in the middle of the day.
        \end{enumerate}
    \end{itemize}
\end{enumerate}

\smallskip

\noindent \textbf{Survey Questions.} Below, we give the questions and the potential answer choices (if they exist) used in our survey for each of the three stages of the survey: 
 
\begin{enumerate}
\item \textbf{Reasoning about an individual member from the lens of  the same individual member}

\emph{Instructions: } You are $X$. You received the following meeting invite from John Doe. Please fill out the following questions from the perspective of $X$.

\emph{Participant Groups:} $X$ can either be Norma Fisher (for Group A), Elizabeth Woods (for Group B), or Theodore Mcgrath (for Group C).

\emph{Questions: }

\begin{enumerate}
    \item To what degree do you believe that the system takes your preferences into account when presenting options? (5-point Likert Scale)
    
    \emph{Choices: Very good (5), Good (4), Average (3), Poor (2), Very Poor (1)}
    
    \item Do you find the reasons behind the options acceptable based on your preferences? (5-point Likert Scale)

    \emph{Choices: Strongly agree (5), Agree (4), Neutral (3), Disagree (2), Strongly Disagree (1)}

    \item If you find the reasoning acceptable, what specific reasons do you find helpful to understand why the options may work? If the reasoning is not acceptable, what type of reasons do you need to understand the options? (Free-text)
\end{enumerate}
\item \textbf{Reasoning about the group from the lens of an individual member}

\emph{Instructions: } Now you get to see everyone's preferences. Please fill out the following questions from the perspective of $X$.

\emph{Participant Groups:} $X$ can either be Norma Fisher (for Group A), Elizabeth Woods (for Group B), or Theodore Mcgrath (for Group C).

\emph{Questions: }

\begin{enumerate}
    \item After seeing all members' individual preferences, to what degree do you believe that the system takes your preferences into account when presenting options? (5-point Likert Scale)

    \emph{Choices: Very good (5), Good (4), Average (3), Poor (2), Very Poor (1)}
    
    \item After seeing all members' individual preferences, do you find the reasons behind the options acceptable? (5-point Likert Scale)

    \emph{Choices: Strongly agree (5), Agree (4), Neutral (3), Disagree (2), Strongly Disagree (1)}
    
    \item How well do you agree the options presented reflect the preferences of the members? (5-point Likert Scale)

    \emph{Choices: Strongly agree (5), Agree (4), Neutral (3), Disagree (2), Strongly Disagree (1)}

\end{enumerate}

\item \textbf{Opinion about our system for collective decision-making}
\begin{enumerate}
    \item In general, how many options would you like a system to propose to you? (Multiple Choice)

    \emph{Choices: 2 options, 3 options, 4 or more options}
    
    \item Now think about meeting scheduling for a group of 5 members. How many options would you like a system to propose to you? (Multiple Choice)

    \emph{Choices: 2 options, 3 options, 4 or more options}
    
    \item How confident are you in using our system to coordinate members' preferences in meeting scheduling? (Free-text)
    \item Do you have any additional feedback or suggestions for this LLM-based system? (Free-text)  
\end{enumerate}

\end{enumerate}

\section{Administrator Interview Questions}

\begin{enumerate}
    \item What types of events do you typically coordinate (e.g., meetings, team outings)? Which occurs most frequently? 

    \item On average, how much time per week do you spend coordinating events (e.g., emails, gathering responses, scheduling)? 
    What percentage of your workweek is this? 

    \item How many events do you typically coordinate in a week? (Include any events you are responsible for, even if ongoing.) 

    \item How many participants do you typically communicate with per event? 

    \item What percentage of participants are from external organizations on average?

    \item Do you find any challenges when coordinating with external participants? Please briefly share the difference between and coordinating completely within your organization. 

    \item What tools do you use for event coordination? Briefly describe your experience, listing the pros and cons. 

    \item What factors do you consider when coordinating events (e.g., power dynamics, location)? Please list them. 

    \item How do you balance these factors? What is your decision-making process? 

    \item After gathering participants' input, how many options do you typically suggest? 

    \item How do you handle situations where no option works for all participants? Briefly describe your approach. 

    \item On average, how many times do you propose new options for an event? 

    \item How often does the final plan not work for everyone? 

    \item What percentage of the initial participants typically agree to or attend the event?

    \item How confident are you that participants' preferences are considered equitably when coordinating events?

    \item What challenges or frustrations do you face during event coordination? What assistance would improve this process? 
\end{enumerate}

\newpage






\end{document}